\documentclass[10pt,journal,compsoc]{IEEEtran}
\ifCLASSOPTIONcompsoc
  \usepackage[nocompress]{cite}
\else
  \usepackage{cite}
\fi
\ifCLASSINFOpdf
\else
\fi
\usepackage{xcolor}
\usepackage{graphicx}
\usepackage{braket,amsfonts,amsopn}
\usepackage[caption=false]{subfig}
\usepackage[utf8]{inputenc}
\usepackage[english]{babel}
\usepackage{amssymb,amsmath}

\usepackage{xspace}
\usepackage{comment}
\usepackage{adjustbox}
\usepackage{enumitem}
\graphicspath{{figs/}}

\usepackage{mathtools}

\usepackage{url}

\usepackage{multirow}

\usepackage{tikz}

\newcommand{\etal}{\emph{et al.}\xspace}
\newcommand{\ie}{\emph{i.e.}\xspace}
\newcommand{\eg}{\emph{e.g.}\xspace}
\newcommand{\wrt}{\emph{w.r.t.}\xspace}
\newcommand{\abbr}{\emph{abbr.}\xspace}

\usepackage{xcolor}

\usepackage{booktabs}

\usepackage{amsmath,mathtools,bm}
\usepackage{geometry}
\newtheorem{myDef}{Definition}

\usepackage{xcolor}

\usepackage{cuted}
\usepackage{flushend}

\usepackage{multicol}
\usepackage[linesnumbered,ruled,vlined]{algorithm2e}

\begin{document}
%
\title{Variational Co-embedding Learning for Attributed Network Clustering}
%
%
%
%

\author{

Shuiqiao Yang,
Sunny Verma, 
Borui Cai,
Jiaojiao Jiang,
Kun Yu, 
Fang Chen,~\IEEEmembership{Senior Member, IEEE}
and~Shui~Yu,~\IEEEmembership{Senior Member, IEEE}
\IEEEcompsocitemizethanks{
\IEEEcompsocthanksitem S. Yang, S. Verma, K. Yu and F. Chen are with Data Science Institute, University of Technology Sydney, Ultimo NSW 2007, Australia. E-mails: shuiqiao.yang@uts.edu.au, sunny.verma@uts.edu.au, kun.yu@uts.edu.au, fang.chen@uts.edu.au.
\IEEEcompsocthanksitem B. Cai is with School of Information Technology, Deakin University, Burwood VIC 3125,  Australia. E-mail: b.cai@deakin.edu.au.
\IEEEcompsocthanksitem J.~Jiang is with School of Computer Science and Engineering, University of New South Wales, Sydney NSW 2052, Australia. E-mail: jiaojiao.jiang@unsw.edu.au.
\IEEEcompsocthanksitem S. Yu is  with the School of Computer Science, University of Technology Sydney, Ultimo NSW 2007, Australia. E-mail: shui.yu@uts.edu.au.
}
\thanks{Manuscript is under review.}
}

\IEEEtitleabstractindextext{%

\begin{abstract}



Recent works for attributed network clustering utilize graph convolution to obtain node embeddings and simultaneously perform clustering assignments on the  embedding space.
It is effective since graph convolution  combines the  structural and attributive information for node embedding learning.
However, a major limitation of such works is that the graph convolution only incorporates the attribute information from the local neighborhood of nodes but fails to exploit the mutual affinities between nodes and attributes.
In this regard, we propose a variational co-embedding learning model for  attributed network clustering (VCLANC). VCLANC is composed of dual variational auto-encoders to simultaneously embed nodes and attributes. 
Relying on this, the mutual affinity information between nodes and attributes could be reconstructed from the embedding space and served as extra self-supervised knowledge for representation learning.
At the same time, trainable Gaussian mixture model  is used as priors  to infer the node clustering assignments.
To strengthen the performance of the inferred clusters, we use a mutual distance loss on the centers of the Gaussian priors and a  clustering assignment hardening loss on the node embeddings. 
Experimental results on four real-world attributed network datasets demonstrate the effectiveness of the proposed VCLANC for attributed network clustering.


\end{abstract}

}

\maketitle

\IEEEdisplaynontitleabstractindextext

%
\IEEEpeerreviewmaketitle



\section{Introduction}


Finding accurate communities or clusters in an attributed network is critical to understand the complex network structures for many downstream applications like group recommendation, user-targeted online advertising, and disease protein discovery \cite{cheng2011clustering,agrawal2018large,8807225}. 
Though  the clustering for attributed network brings many important applications, but it also poses significant new challenges.
Firstly, as the attributed network includes not only structural connections but also attribute values, it is difficult to naturally combine and leverage the two types of information in the process of clustering \cite{xusigmoid12}.
Furthermore, it is usually hard to find supervision information to guide the cluster discovery in attributed network \cite{sun2020network,bo2020www,WangPHLJZ19,zhang2019attributed}. 

To handle the challenges,  many network embedding and graph neural network related methods have been developed recently for node representation learning to improve the accuracy of downstream applications like graph classification, link prediction and  graph clustering.
The representative  methods include DeepWalk \cite{deepwalk2014Perozzi}, Line \cite{line2015Tang}, node2vec \cite{node2vec2016Grover}, struc2vec \cite{struc2vec2017KDD}, GCN \cite{kipf2016semi}, GraphSAGE \cite{graphsage2017Hamilton}, ARGA/ARVGA\cite{pan2018adversarially}, MGAE\cite{wang2017mgae}, AGC \cite{zhang2019attributed}, etc.
For example, DeepWalk \cite{deepwalk2014Perozzi} and Node2Vec \cite{node2vec2016Grover} are two representative network embedding methods to learn low-dimensional representations for nodes through the local neighborhood structure prediction. 
Then, general clustering algorithms such as $K$-means can be adopted on the learned node embeddings  for the network clustering task.
Similarly, the graph convolutional network (GCN)  proposed by Kipf \etal has also been widely exploited into the graph clustering task \cite{kipf2016semi}. 
For example, AGC  exploits a $K$-order GCN to update the features of each node via the  $k$-hop neighbors and then adopts spectral clustering to find the best node partitions based on the new features for each node.
These decoupled clustering strategies show great improvement in network clustering due to the improved node representations. 
However, they are sub-optimal in nature as  the node representation learning process are independent with the node clustering process.

Later, graph auto-encoder based framework with self-contained clustering strategies have been proposed for attributed network clustering to overcome the aforementioned drawbacks. 
The  methods include VGAECD \cite{Choong2018}, DAEGC \cite{WangPHLJZ19},  NEC \cite{sun2020network}, SDCN \cite{bo2020www}, etc. 
In these methods, graph neural network is used as the encoder and inner product between latent variables would be the decoder to reconstruct the input data for self-supervised learning \cite{kipf2016variational}.
A clustering objective is jointly optimized with the  reconstruction loss to learn node embeddings that are optimal for clustering purpose.
For example, VGAECD \cite{Choong2018} finds the community structure by jointly optimizing a variational auto-encoder (VAE) and a Gaussian mixture model. 
NEC  \cite{sun2020network} jointly learns the node embedding and node community assignments by adding a soft clustering loss and modularity loss into the graph auto-encoder framework.
SDCN \cite{bo2020www} integrates the structural information for clustering  using a DNN module and a GCN module to learn different aspects of information, and adopts a dual self-supervised mechanism to unify the learned representation from the two modules.

However, these methods only aggregate the attributes from the local neighborhood for node representation learning but ignore the mutual affinity information between nodes and attributes. 
But mutual affinities between nodes and attributes such that a node contains different attributes and the same attribute could be shared by different nodes can denote much deep and implicit information. 
For example, much fine-grained node similarities at attribute level could be retained given the affinities from nodes to attributes.
Therefore, exploiting this affinities should be useful for learning better node representations.
Moreover, these methods may not be able to learn more distinguishable representations due to the lack of separate prior distributions for different node clusters to regularize the node embedding learning.

To alleviate the aforementioned drawbacks, we propose a  \textbf{V}ariational \textbf{C}o-embedding \textbf{L}earning model for \textbf{A}ttributed \textbf{N}etwork \textbf{C}lustering, abbreviated as VCLANC, which aims to embed the nodes and  attributes into the same semantic space where the mutual affinities between nodes and attributes could be reconstructed for better representation learning.
At the same time, we exploit a trainable Gaussian mixture model as priors on the latent variables to learn clustering friendly node embeddings and jointly infer the nodes clustering assignments.
A mutual distance loss is deployed on the Gaussian priors to force different priors to become more separable.
Also, a  clustering assignment hardening loss is jointly optimized in the co-embedding learning process to further strength the clustering assignment qualities for the nodes.

Our contributions are summarized as follows:

\renewcommand{\labelenumi}{\Roman{enumi}.}

\begin{enumerate}
    \item We propose a variational co-embedding learning model named VCLANC for attributed network clustering.
    VCLANC exploits the mutual affinities between nodes and attributes as extra self-supervised knowledge to learn  better node representations by using dual variational auto-encoders to co-embed  nodes and attributes in the same latent space. 
    \item 
    We simultaneously optimize trainable Gaussian mixture model with the dual variational auto-encoders in VCLANC to infer the clustering assignments in the embedding space.
    Meanwhile, we further adopt a mutual distance loss on Gaussian priors and a clustering assignment hardening loss on the node embedding space to strengthen the clustering effectiveness.
	\item We conduct  experiments on four real-world attributed network datasets to verify the effectiveness of VCLANC. The experimental results demonstrate the  proposed method is more effective in the network clustering task and node embedding learning than the state-of-the-art methods.
\end{enumerate}

The rest of this paper is organized as follows.
Section 2 presents the related work. 
Section 3 introduces the notations, problem definition and preliminary.
Section 4 details the proposed approach. 
Experimental setup is introduced in section 5.
Experimental results are reported in section 6.
Conclusions are made in section 7. 

\section{Related Work}
We classify the related work for network clustering into two groups. 
The first group  includes the traditional community detection related methods. The second group includes the deep learning based methods.
\subsection{Community detection}

For the non-attributed network, various methods have been developed to exploit the link structure of the networks to derive different connective patterns and partition the network into different communities.  
For example, the graph partitioning based methods iteratively divide the network until the stopping criteria is reached. 
For example, Kernighan \etal \cite{kernighan1970efficient} have proposed  a heuristic algorithm called Kernighan–Lin to randomly divide the network into two subsets and exchange the nodes between the subsets to find the optimal communities.
The spectral clustering based methods have also been developed for community detection. 
For example, Donath \etal \cite{donath2003lower} have proposed to use the largest eigenvalues of the adjacent matrix for community detection.
Shi \etal \cite{shi2000normalized} have proposed the normalized spectral clustering  method for community detection.
Also, the  modularity based methods have been proposed to find the optimal communities in network by maximizing the graph modularity \cite{newman2006modularity,shang2013community}. 
For example, Newman \etal \cite{newman2006modularity} claim that the network modularity could be expressed as eigenvectors of a  characteristic matrix. They first construct the modularity matrix for the target network and find the most positive eigenvalue with the associated eigenvector. Then, they divide the network into two parts based on the sighs of the elements in eigenvectors and repeat this process till the sub-networks are indivisible. 
Shang \etal \cite{shang2013community} have proposed MIGA based on modularity and an improved genetic algorithm for community detection.
They exploit the modularity as the objective function and the prior information of the community structures as target to make the community detection become more stable and accurate. 
The matrix factorization  based methods focus on learning the main components to recover the adjacency matrix of a given network and thus reveal the network communities.
Yang \etal \cite{yang2013overlapping} have proposed a method called BIGCLAM to find the overlapping communities through nonnegative matrix factorization. 

Compared with the structure only non-attributed networks, the attributed networks with node features contain more comprehensive information.
Hence, community detection in attributed networks has attracted much attention in recent years. 
For example, Neville \etal \cite{neville2003clustering} define distance measurement between attributed nodes to weight the edges for better community detection.
After that, the classical methods such as Karger's Min-Cut \cite{karger1993global} and spectral clustering \cite{ng2002spectral} are exploited on the weighted adjacent matrix to community detection or node clustering.
Yang and Hong \etal \cite{zhou2010clustering,cheng2011clustering} have proposed SA-Cluster which first  computes an augmented graph computed from the  attribute information of nodes. Then, SA-Cluster exploits neighborhood random walk to compute a unified distance between vertices on using both the raw topological structure and augmented graph for community detection or node clustering. 
Combe \cite{combe2012combining} have adopted a hierarchical agglomerative clustering method based on a defined unified cosine distance on textual information of the nodes and geodesic distance on the network structure for community detection.
Xu \etal \cite{xusigmoid12} have proposed to integrate  the structural and attribute information into a  Bayesian model for clustering an attributed graph. In the Bayesian model, the clustering problem is transferred into a standard probabilistic inference problem, which is solved by a variational algorithm.

\subsection{Deep learning for network clustering}
To our best understanding, the methods for graph clustering via network embeddings or graph neural networks could be categorized into two types based on the different clustering strategies.
In the first type, the methods focus on learning better representations for nodes and then general clustering method like $K$-means is adopted for network clustering.
For example, DeepWalk \cite{deepwalk2014Perozzi} learns the embedding for nodes through maximizing the probability of predicting its closest neighbors. 
The neighbors of a node are derived through random walk on the network structure. 
Node2Vec \cite{node2vec2016Grover} make improvements  for the node representation learning by optimizing a neighborhood preserving objective. 
The struc2vec \cite{struc2vec2017KDD} learns the node representation through a multilayer graph to encode structural similarities and structural context for nodes.
Based on the improved representations of nodes, better network clustering performance could be reached using general clustering methods.

Later, graph convolutional network (GCN) is proposed and quickly becomes popular for graph data processing \cite{kipf2016semi}. 
GCN creatively devises a graph convolution operation to aggregate the information from local neighborhood for node representation learning.
Many variants have been proposed based on GCN.
For example, GraphSAGE \cite{graphsage2017Hamilton} improves GCN with trainable aggregator functions for better feature  aggregation from the local neighborhood of nodes.
Also, graph auto-encoder (GAE) and varitional graph auto-encoder (VGAE) have been proposed which using the GCN as encoder and the inner product between node embeddings as decoder in auto-encoder or variational auto-encoder framework for unsupervised representation learning \cite{kingma2013auto, kipf2016variational}.
Based on GAE and VGAE, Pan \etal \cite{pan2018adversarially} have proposed ARGA and ARVGA with adversarial training scheme where the latent node representations are enforced to match a prior distribution.
Wang \etal \cite{wang2017mgae} have proposed MGAE which is a GAE with a  marginalization process to disturb the network attribute information.

The second group of methods for graph clustering aim to jointly infer the clustering assignments and node embeddings. 
For example, Wang \etal \cite{WangPHLJZ19} have proposed DAEGC, which exploits attentional strategy into GAE to determine the importance of neighboring nodes and uses a soft clustering loss as clustering objective to jointly infer the node clustering assignments.
Sun \etal \cite{sun2020network} have proposed NEC for attributed graph clustering using GAE with the same clustering objective in used DAEGC \cite{WangPHLJZ19}. 
Additionally, NEC introduces a relaxed modularity loss to  exploit the higher-order proximity of nodes for preserving community structure in the node representation learning. 
Bo \etal \cite{bo2020www} have proposed SDCN which combines a DNN module, a GCN module, and a dual self-supervised module for comprehensive representation learning and graph clustering. 
Similarly, Li \etal \cite{li2020deep} also have proposed DGSCN which combines both graph node feature and topology information with  deep GCN layers and  a triple self-supervised module for graph clustering.
Choong \etal \cite{Choong2018} have proposed VGAECD for attributed graph clustering. Unlike DAEGC and NEC, VGAECD exploits VGAE with a  Gaussian mixture model based generative process to infer the clustering assignments.

\section{Notations, Problem Definition and Preliminary}

In this section, we introduce the mainly used notations and  formally define the studied problem. After that, we introduce the preliminary for variational auto-encoder.
The terms network and graph will be used interchangeably.

\subsection{Notations}

\begin{table}[t]
    \centering
    \caption{Notations used in this paper.}
    \begin{tabular}{ll}
    \toprule
    Notation & Description \\
    \toprule
    $\mathcal{G}$ & attributed network \\
    $\mathcal{V}$ & node set \\
    $\mathcal{A}$ & attribute set \\
    $\mathbf{A} \in \mathbb{R}^{N \times N}$ & adjacency matrix for network \\
    $\mathbf{X} \in \mathbb{R}^{N \times M}$ & attribute matrix for nodes \\
    $N=|\mathcal{V}|$ & number of nodes \\
    $M=|\mathcal{A}|$ & number of attributes \\
    $J$ & dimension of latent variables \\
    
    $\mathbf{Z}^{\mathcal{V}} \in \mathbb{R}^{N \times J}$ & embedding matrix for the nodes \\
    $\mathbf{Z}^{\mathcal{A}} \in \mathbb{R}^{M \times J}$ & embedding matrix for the attributes \\
    \bottomrule
    \end{tabular}
     \label{tab:notations}
\end{table}


Table \ref{tab:notations} summarizes the mainly adopted notations in this paper. 
An attributed network is denoted as $\mathcal{G}$ with four elements: the node set $\mathcal{V}$, the attribute set $\mathcal{A}$, the adjacent matrix $\mathbf{A}$ and the attribute matrix $\mathbf{X}$.
$\mathcal{V}$ represents all the nodes in graph $\mathcal{G}$ and its size is  $N=|\mathcal{V}|$.
The structural connections of $\mathcal{G}$ are denoted with an adjacent matrix $\mathbf{A} \in \mathbb{R}^{N\times N}$ that is binary-valued (\ie, zero or one ) to represent the  existence of edges between nodes, \eg, $\mathbf{A}_{ij}=1$ denotes that node $i$ and node $j$ have connection.
The attributes of nodes in graph $\mathcal{G}$ are denoted as a matrix $\mathbf{X} \in \mathbb{R}^{N\times M}$, where the $i^{th}$ row of $\mathbf{X}$  is represented with binary values which indicate the appearance of the raw attributes for $i^{th}$ node in dimension size of $M$.
Meanwhile, the $j^{th}$ column of $\mathbf{X}$ denotes how the $j^{th}$ attribute is distributed among different nodes. 
$\mathbf{A}$ represents the network structure and $\mathbf{X}$ represents the mutual affinities between nodes and attributes.
The embeddings for nodes and attributes are denoted as $\mathbf{Z^{\mathcal{V}}} \in \mathbb{R}^{N \times J}$ and $\mathbf{Z^{\mathcal{A}}} \in \mathbb{R}^{N \times J}$, where the dimension size for the embedding space is $J$.

\subsection{Problem Definition}

With the adopted notations introduced above, we formally define the studied problem of attributed network clustering. 

\begin{myDef}[Attributed network clustering]\label{def:problem}
For a given attributed network $\mathcal{G}=(\mathcal{V},\mathcal{A},\mathbf{A},\mathbf{X})$, the clustering for an attributed network is to find a partition function $f$ that exploits  the attribute matrix $\mathbf{X}$ and adjacent matrix $\mathbf{A}$ of $\mathcal{G}$ as inputs to divide the nodes $\mathcal{V}$ into disjoint sets in an unsupervised manner as follows:
\begin{equation}
    (\mathbf{A},\mathbf{X})\xrightarrow{f} (\hat{\mathcal{V}}_1, \hat{\mathcal{V}}_2, \cdots, \hat{\mathcal{V}}_k),  
\end{equation}
such that: (1) the nodes with similar graph structure are more likely to be partitioned into the same set; (2) the nodes within a set are more likely to share similar attributes.
\end{myDef}


\subsection{Preliminary for Variational Auto-encoder}

\begin{figure}[t]
    \centering
    \includegraphics[width=0.4\textwidth]{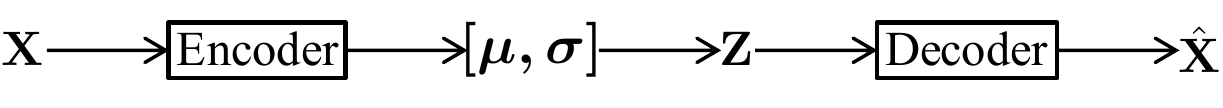}
    \caption{The framework of variational auto-encoder (VAE). 
    Its encoder is an inference network which learns the latent representation for the inputs as Gaussian distributions.
    Its decoder is a generative network to reconstruct the input data.}
    \label{fig:vae_framework}
\end{figure}

The variational auto-encoder (VAE) is a self-supervised representation learning model 
which approximates the posterior distribution of latent variable  and maximizes the marginal likelihood of the input data with an evidence lower-bound (\textit{ELBO})  \cite{kingma2013auto}.
Fig. \ref{fig:vae_framework} shows the framework of the VAE model.
For a given latent variable model  $p_{\boldsymbol{\theta}}(\mathbf{X,Z}) = p_{\boldsymbol{\theta}}(\mathbf{X}|\mathbf{Z})p(\mathbf{Z})$ where $\mathbf{X}$ denotes the observed data and $\mathbf{Z}$ denotes the latent representation, VAE uses an approximated posterior distribution $q_{{\boldsymbol{{\phi}}}}(\mathbf{Z}|\mathbf{X})$ for the intractable true posterior and then learns the optimal variational parameter ${\boldsymbol{\phi}}$ and the generative model parameters $\boldsymbol{\theta}$ through maximizing an evidence lower bound (\textit{ELBO}) defined as follows:

\begin{equation}
\label{eq:vae_elbo}
\begin{aligned}
\mathcal{L}_{e}(\boldsymbol{\theta},\boldsymbol{{\boldsymbol{{\phi}}}};\mathbf{X})  = &  \mathbb{E}_{q_{\boldsymbol{\phi}}}[\log p_{\boldsymbol{\theta}}(\mathbf{X}|\mathbf{Z})]- D_{KL}(q_{{\boldsymbol{{\phi}}}}(\mathbf{Z}|\mathbf{X})\ \| \ p(\mathbf{Z})) \\
    & \leq \log p(\mathbf{X}), 
\end{aligned}
\end{equation}

where the  posterior approximation distribution is generally a simple distribution such as a multivariate Gaussian distribution with diagonal covariance. $D_{KL}(q_{{\boldsymbol{{\phi}}}}(\mathbf{Z}|\mathbf{X})\ \| \ p(\mathbf{Z}))$ is the Kullback-Leibler divergence (KL divergence) to measure the similarity between two probability distributions and it acts as a regularizer to force the  posterior approximation distribution $q_{{\boldsymbol{{\phi}}}}(\mathbf{Z}|\mathbf{X})$ to be closer to the prior $p(\mathbf{Z})$.

As shown in Fig. \ref{fig:vae_framework}, VAE is consisted with two computational networks: an \textit{inference network}  $f_{\boldsymbol{{\boldsymbol{{\phi}}}}}(\cdot)$ called encoder and a \textit{generative network}  $g_{\boldsymbol{\theta}}(\cdot)$ called decoder.  
The inference network $f_{{\boldsymbol{{\phi}}}}(\cdot)$ is used to simulate the latent variable distribution $q_{{\boldsymbol{{\phi}}}}(\mathbf{Z}|\mathbf{X})$  while the  generative network $g_{\boldsymbol{\theta}}$ is used to simulate $p_{\boldsymbol{\theta}}(\mathbf{X}|\mathbf{Z})$ to reversely generate the input data  from the latent variables $\mathbf{Z}$.
To learn the optimal representations for the raw data, VAE updates  $f_{\boldsymbol{{\boldsymbol{{\phi}}}}}(\cdot)$ and $g_{\boldsymbol{\theta}}(\cdot)$ simultaneously through optimizing the \textit{ELBO} in Eq. \ref{eq:vae_elbo}.
To generalize the training process of VAE, a reparameterization trick is adopted to exploit the general optimization methods like stochastic gradient descent\cite{kingma2013auto}.
The structures for $f_{\boldsymbol{{\boldsymbol{{\phi}}}}}(\cdot)$ and $g_{\boldsymbol{\theta}}(\cdot)$ could be chose flexibly given different tasks. 
For instance, variational graph neural network (VGAE)  \cite{kipf2016variational} has been proposed to learn the embeddings for graph-structured data using graph convolutional neural network (GCN) \cite{kipf2016semi} as the encoder.

\section{Proposed method}

In this section, we elaborate  our proposed method: variational co-embedding learning for attributed network clustering (VCLANC). 

\subsection{Overview of VCLANC}

\begin{figure}[t]
    \centering
    \includegraphics[width=0.5\textwidth]{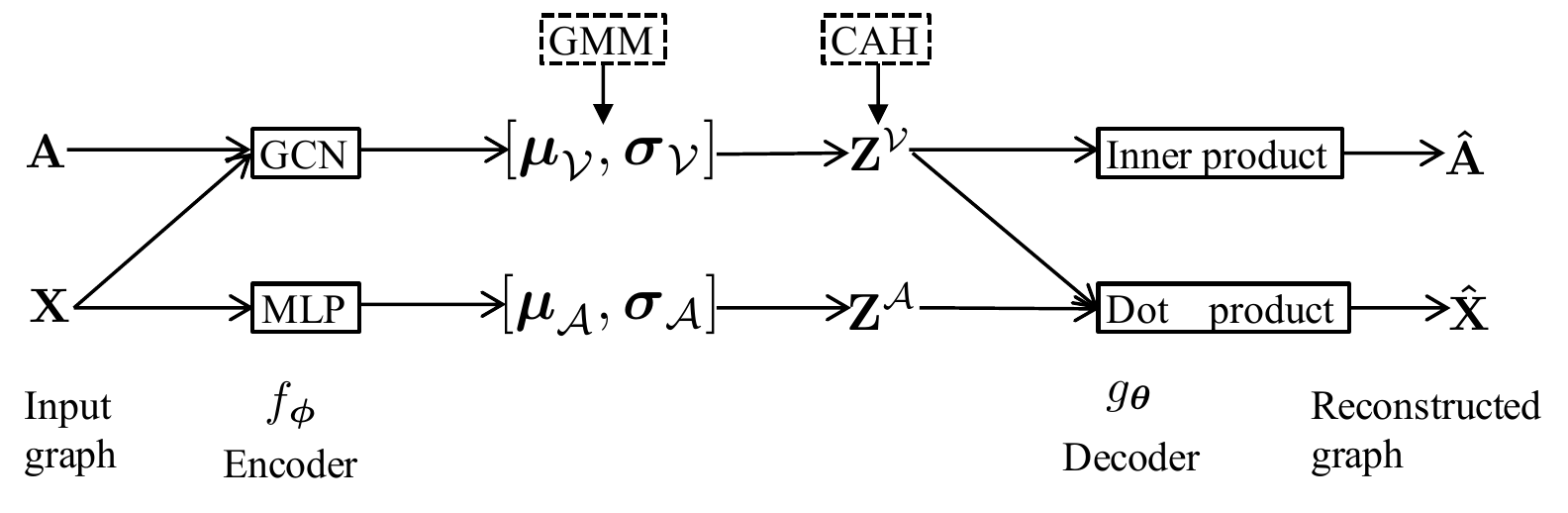}
    \caption{The structure of the proposed VCLANC model. 
    A two-layer GCN and two-layer multilayer perceptron (MLP)  are used as encoders to embed both  nodes and attributes as Gaussian distributions with means and  variances.
    Then, simple inner and dot product is used as encoders to reconstruct the input data.
    Trainable  Gaussian mixture model (GMM) is used to infer the node clustering assignments  and clustering assignment hardening (CAH) loss is adopted on node embeddings space to strengthen clustering quality.}
    \label{fig:framework}
\end{figure}

Fig. \ref{fig:framework} shows the overall structure of VCLANC.
VCLANC adopts dual variational auto-encoders to co-embed the nodes and attributes into the same latent space.
A two-layer graph neural network (GCN) is used as an encoder to fuse the network topology $\mathbf{A}$ and the  attribute matrix $\mathbf{X}$ for learning  node embeddings  $\mathbf{Z}^{\mathcal{V}}$.
The inner product among $\mathbf{Z}^{\mathcal{V}}$ is used as the decoder to reconstruct the adjacent matrix as $\mathbf{\hat{A}}$.
To exploit  mutual affinities between nodes and attributes, VCLANC uses a  two-layer multilayer perceptron (MLP) as encoder to embed the attributes of the network into the same latent space with nodes as $\mathbf{Z}^{\mathcal{A}}$ and then adopts the dot product between  $\mathbf{Z}^{\mathcal{V}}$ and  $\mathbf{Z}^{\mathcal{A}}$ as decoder to reconstruct the attribute matrix $\mathbf{X}$.
The reconstructed $\mathbf{\hat{X}}$ and $\mathbf{\hat{A}}$ are used as self-supervised knowledge for optimizing the parameters of the dual variational auto-encoders.

To infer node clusters, we assume that the latent node embeddings of different groups follows Gaussian distributions as their priors.
We incorporate the Gaussian mixture model (GMM) as priors to infer cluster assignments in the latent space.
Meanwhile, a clustering assignment hardening (CAH) loss which adopts KL divergence to strengthen the confident nodes assignments from GMM is adopted to 
ensure better clustering qualities.
We detail  VCLANC in the following sections.

\subsection{Generative Process for Attributed Graph with GMM}

In our method, we use a  GMM as priors to guide the embedding learning  and infer the nodes partitions for an attributed graph.
The formulation of an attributed graph with $\mathbf{A} $ and $\mathbf{X}$ could be explained through a generative process as follows:

 \begin{equation}\label{eq:pi}
 \begin{array}{l}
      p(\boldsymbol{c};\boldsymbol{\pi}) = \operatorname{Cat}(\boldsymbol{\pi}) \\ 
      p(\boldsymbol{z}_i \ \mid\  \boldsymbol{c}_{k}=1)=\mathcal{N}(\boldsymbol{\mu}_{k}, \mathbf{I}(\boldsymbol{\sigma}_{k}^{2})) \\  
      
      [\boldsymbol{\mu}_{z},  \boldsymbol{\sigma}_{z}] = f_{\boldsymbol{\theta}}(\mathbf{z}_i) \\ 
      
    p_{\boldsymbol{}}(\mathbf{A}_i \ \mid\  \boldsymbol{z}_i)=\left\{
    \begin{array}{ll}
    \operatorname{Ber}\left(\boldsymbol{\mu}_{z}\right) & ~~~~~~~~~~~~~~\text { if } \mathbf{A}_i \in \{0,1\}^N \\
    \mathcal{N}\left(\boldsymbol{\mu}_{z},  \mathbf{I}\boldsymbol{\sigma}_{z}^{2}\right) & ~~~~~~~~~~~~~~\text { if } \mathbf{A}_i \in \mathbb{R^N}
\end{array}\right.
 \end{array}
 \end{equation}

 
 
\begin{equation}\label{eq:atrribute_embedding}
\begin{array}{l}
p_{\boldsymbol{\theta}}(\boldsymbol{a}_j)=\mathcal{N}(0, \mathbf{I}) \\ 
 
[\boldsymbol{\mu}_{a}, \boldsymbol{\sigma}_{a}] = f_{\boldsymbol{\theta}}(\mathbf{a}_j) \\ 

p_{\boldsymbol{\theta}}(\mathbf{X}_{ij}|\boldsymbol{z}_i, \boldsymbol{a}_j)  
=   \left\{

\begin{array}{ll}
\operatorname{Ber}\left(\boldsymbol{\mu}_{a}^{T}\boldsymbol{\mu}_{z}\right) & \text { if } \mathbf{X}_{ij} \in \{0,1\} \\
\mathcal{N}\left(\boldsymbol{\mu}_{a}^{T}\boldsymbol{\mu}_{z}, \mathbf{I}\boldsymbol{\sigma}_{a}^2\right) & \text { if } \mathbf{X}_{ij} \in \mathbb{R}
\end{array}\right.

\end{array}
\end{equation}

Eq. \ref{eq:pi} shows the generative process network structure, where $\boldsymbol{z}_i \in \mathbf{Z}^\mathcal{V}$ and $\mathbf{A}_i$ are the embedding and neighborhood structure of the $i^{th}$ node, respectively.  
The parameters for GMM priors are $\{\boldsymbol{\mu}, \boldsymbol{\sigma},\boldsymbol{\pi}\}$. 
$f_{\boldsymbol{\theta}}(\cdot)$ is a neural network to learn the expectation of parameters.
To generate the neighborhood structure for a node, we first need to sample an indicator vector $\boldsymbol{c}\in \{0,1\}^{K}$ from a categorical distribution $\operatorname{Cat}(\boldsymbol{\pi})$ to determine which Gaussian prior distribution would be used to generate the latent representation $\boldsymbol{z}_i$.
Then, the network structure $\mathbf{A}_i$ could be generated from either multivariate Bernoulli or Gaussian distributions. The parameters $\boldsymbol{\mu}_x$ or $\boldsymbol{\sigma}_x$ are generated by a neural network $f_{\boldsymbol{\theta}}(\cdot)$.
Similarly, Eq. \ref{eq:atrribute_embedding} shows the generative process for attribute except that we assume that the latent representations of attributes follow a standard multivariate Gaussian distribution. 

\subsection{Evidence Lower Bound of VCLANC} 



Based on the generative process for an attribute graph, the joint probability distribution between the latent variables  $(\boldsymbol{Z}^{\mathcal{V}}$, $\boldsymbol{Z}^{\mathcal{A}}$, $\mathbf{C})$ and the observed variables $(\mathbf{A},\mathbf{X})$ could be represented as follows:

\begin{equation}\label{eq: joint_probability}
\begin{array}{l} 

 p_{\boldsymbol{\theta}}(\mathbf{A},\mathbf{X} ,\mathbf{Z}^{\mathcal{V}}, \mathbf{Z}^{\mathcal{A}}, \mathbf{C}) \\ 
= p(\mathbf{C})p(\mathbf{Z}^{\mathcal{V}}\ |\ \mathbf{C}) p(\mathbf{Z}^\mathcal{A}) p_{\boldsymbol{\theta}_1}(\mathbf{A}\ |\ \mathbf{Z}^\mathcal{V})p_{\boldsymbol{\theta}_2}(\mathbf{X}\ |\ \mathbf{Z}^\mathcal{A},\mathbf{Z}^{\mathcal{V}}).
\end{array}
\end{equation}

Then, according to Jensen's inequality, the \textit{ELBO} of $\log p(\mathbf{A},\mathbf{X})$ can be derived as follows:

\begin{equation}\label{eq:elbo}
    \begin{aligned}
    \begin{array}{l}
         \log p(\mathbf{A},\mathbf{X}) \\ 
         = \log \int_{\mathbf{Z}^{\mathcal{A}}} \int_{\mathbf{Z}^{\mathcal{V}}}\int_{\mathbf{C}} p_{\boldsymbol{\theta}}(\mathbf{A},\mathbf{X}, \mathbf{Z}^{\mathcal{A}},\mathbf{Z}^{\mathcal{V}},\mathbf{C}) d_{\mathbf{Z}^{\mathcal{A}}}d_{\mathbf{Z}^{\mathcal{V}}}d_{\mathbf{C}}\\
         = \log \int_{\mathbf{Z}^{\mathcal{A}}} \int_{\mathbf{Z}^{\mathcal{V}}}\int_{\mathbf{C}} \dfrac{p_{\boldsymbol{\theta}}(\mathbf{A},\mathbf{X}, \mathbf{Z}^{\mathcal{A}},\mathbf{Z}^{\mathcal{V}},\mathbf{C}) q_{\boldsymbol{{\phi}}}(\mathbf{Z}^{\mathcal{A}},\mathbf{Z}^{\mathcal{V}},\mathbf{C}\ |\ \mathbf{A},\mathbf{X})}{q_{\boldsymbol{{\phi}}}(\mathbf{Z}^{\mathcal{A}},\mathbf{Z}^{\mathcal{V}},\mathbf{C}\ |\ \mathbf{A},\mathbf{X})} \\ d_{\mathbf{Z}^{\mathcal{A}}}d_{\mathbf{Z}^{\mathcal{V}}}d_{\mathbf{C}}  \\
          
         \geq  \mathbb{E}_{ q_{\boldsymbol{{\phi}}}(\mathbf{Z}^{\mathcal{A}},\mathbf{Z}^{\mathcal{V}},\mathbf{C}\ |\ \mathbf{A},\mathbf{X})} \left[\log \dfrac{p_{\boldsymbol{\theta}}(\mathbf{A},\mathbf{X}, \mathbf{Z}^{\mathcal{A}},\mathbf{Z}^{\mathcal{V}},\mathbf{C})}{q_{\boldsymbol{{\phi}}}(\mathbf{Z}^{\mathcal{A}},\mathbf{Z}^{\mathcal{V}},\mathbf{C}\ |\ \mathbf{A},\mathbf{X})}\right],
         \\\overset{\Delta}{=}\mathcal{L}_{e}(\boldsymbol{\theta},{\boldsymbol{{\phi}}};\mathbf{A},\mathbf{X}), 
    \end{array}
    \end{aligned}
\end{equation}





where $\mathcal{L}_e(\boldsymbol{\theta},{\boldsymbol{{\phi}}};\mathbf{A},\mathbf{X})$ is the ELBO for the marginal log-likelihood of the observed variables: $\mathbf{A}$ and $\mathbf{X}$. 
$q_{\boldsymbol{{\phi}}}(\mathbf{Z}^{\mathcal{A}},\mathbf{Z}^{\mathcal{V}},\mathbf{C}\ |\ \mathbf{A},\mathbf{X})$ (abbreviated as $q_{\boldsymbol{{\phi}}}$ for simplicity) is  a simpler distribution used to approximate the intractable true posterior $p(\mathbf{Z}^{\mathcal{A}},\mathbf{Z}^{\mathcal{V}},\mathbf{C}\ |\ \mathbf{A},\mathbf{X})$. 
With the mean-field assumption, the latent variables are independent to each other and thus the  posterior approximation $q_{\boldsymbol{{\phi}}}(\mathbf{Z}^{\mathcal{A}},\mathbf{Z}^{\mathcal{V}},\mathbf{C}\ |\ \mathbf{A},\mathbf{X})$ could be factorized as follows: 
\begin{equation}\label{eq:approximated_posterior_fact}
    \begin{aligned}
    \begin{array}{ll}
    q_{\boldsymbol{{\phi}}}(\mathbf{Z}^{\mathcal{A}},\mathbf{Z}^{\mathcal{V}},\mathbf{C}\ |\ \mathbf{A},\mathbf{X}) & \\ 
    = q_{{\boldsymbol{{\phi}}}_1}(\mathbf{Z}^{\mathcal{A}}\ |\ \mathbf{X} )q_{{\boldsymbol{{\phi}}}_2}(\mathbf{Z}^{\mathcal{V}}\ |\ \mathbf{A},\mathbf{X}) q_{{\boldsymbol{{\phi}}}_3}(\mathbf{C}\ |\ \mathbf{A},\mathbf{X})& \\
    \end{array}
    \end{aligned}
\end{equation}

Therefore, the \textit{ELBO} in Eq. \ref{eq:elbo} can be further decomposed by substituting Eq. \ref{eq: joint_probability} and  \ref{eq:approximated_posterior_fact} as follows:

\begin{equation}\label{eq:elbo_fact}
    \begin{aligned}
    \begin{array}{ll}
         & \mathcal{L}_{e}(\boldsymbol{\theta},{\boldsymbol{{\phi}}};\mathbf{A},\mathbf{X})  \\   
        &=  \mathbb{E}_{q_{\boldsymbol{{\phi}}}}[\log p_{\boldsymbol{\theta}_1}(\mathbf{A}\ |\  \mathbf{Z}^{\mathcal{V}}) ]  \\ 
        &+ \mathbb{E}_{q_{{\boldsymbol{{\phi}}}}}[\log p_{\boldsymbol{\theta}_2}(\mathbf{X}\ |\ \mathbf{Z}^{\mathcal{A}},\mathbf{Z}^{\mathcal{V}})]  \\
        &- D_{KL}(q_{{\boldsymbol{{\phi}}}_1}(\mathbf{Z}^\mathcal{A}\ |\  \mathbf{X})\ \| \ p(\mathbf{Z}^\mathcal{A}))  \\
        &- \sum\limits_{\mathbf{C}}q_{{\boldsymbol{{\phi}}}_3}(\mathbf{C}\ |\ \mathbf{A},\mathbf{X}) D_{KL}(q_{{\boldsymbol{{\phi}}}_2}(\mathbf{Z}^\mathcal{V}\ |\  \mathbf{A},\mathbf{X})\ \| \ p(\mathbf{Z}^{\mathcal{V}}\ |\ \mathbf{C}))  \\ 
        &- D_{KL}(q_{{\boldsymbol{{\phi}}}_3}(\mathbf{C}\ |\ \mathbf{A},\mathbf{X})\ \| \ p(\mathbf{C})).  \\
    \end{array}
    \end{aligned}
\end{equation}

$q_{{\boldsymbol{{\phi}}}_1}(\mathbf{Z}^\mathcal{A}\ |\  \mathbf{X})$ and  $q_{{\boldsymbol{{\phi}}}_2}(\mathbf{Z}^\mathcal{V}\ |\  \mathbf{A},\mathbf{X})$ are  the probabilistic encoders which produce the posterior approximation distributions (\eg, Gaussian) given the input data $\mathbf{A}$ and $\mathbf{X}$. 
The encoders could have flexible structures for different tasks. 
As shown in Fig. \ref{fig:framework}, we use the graph convolutional network (GCN) for $q_{{\boldsymbol{{\phi}}}_2}(\mathbf{Z}^\mathcal{V}\ |\  \mathbf{A},\mathbf{X})$ 
 and the multilayer perceptron (MLP) for   $q_{{\boldsymbol{{\phi}}}_1}(\mathbf{Z}^\mathcal{A}\ |\  \mathbf{X})$, respectively. 
Similarly, $p_{\boldsymbol{\theta}_1}(\mathbf{A}\ |\  \mathbf{Z}^{\mathcal{A}})$ and $p_{\boldsymbol{\theta}_2}(\mathbf{X}\ |\ \mathbf{Z}^{\mathcal{A}},\mathbf{Z}^{\mathcal{V}})$ are the probabilistic decoders that given the latent representations, they would produce distribution for the observed data $\mathbf{A}$ and $\mathbf{X}$.

Following VAE \cite{kingma2013auto}, the priors and posterior approximations for the embeddings are set to be multivariate Gaussian distributions with diagonal covariance as follows: 

\begin{equation}\label{eq:prior_postetior_parameters}
    \begin{array}{l}
    p(\mathbf{Z}_i^\mathcal{V}|c) \sim \mathcal{N}(\boldsymbol{\hat{\mu}}_{c},\mathbf{I}\boldsymbol{\hat{\sigma}}_{c}^2) \\
    p(\mathbf{Z}^{\mathcal{A}}) = \mathcal{N}(0,\mathbf{I}\boldsymbol{\hat{\sigma}}^2_{\mathcal{A}}) \\ 
    q_{{\boldsymbol{{\phi}}}_1}(\mathbf{Z}_i^\mathcal{A}|\mathbf{X}) = \mathcal{N}(\boldsymbol{\mu}_{\mathcal{A}_i},\mathbf{I}\boldsymbol{\sigma}_{\mathcal{A}_i}^2)  \\
    q_{{\boldsymbol{{\phi}}}_2}(\mathbf{Z}_i^\mathcal{V}|\mathbf{A},\mathbf{X}) =  \mathcal{N}(\boldsymbol{\mu}_{\mathcal{V}_i},\mathbf{I}\boldsymbol{\sigma}_{\mathcal{V}_i}^2).  \\ 
    \end{array}
\end{equation}

$\boldsymbol{\hat{\mu}}_{c}$ and $\boldsymbol{\hat{\sigma}}_c^2$ represent one of the Gaussian priors for node embedding distribution.
We assume a normal Gaussian prior for attribute embeddings with mean and diagonal covariance as 0 and $\mathbf{I} \boldsymbol{\hat{\sigma}}^2_{\mathcal{A}}$, respectively. 
$\boldsymbol{\mu}_{\mathcal{V}_i}$,$\boldsymbol{\mu}_{\mathcal{A}_i}$,$\boldsymbol{\sigma}_{\mathcal{V}_i}^2$ and $\boldsymbol{\sigma}_{\mathcal{A}_i}^2$ are the parameters of Gaussian distributions for nodes and attributes embeddings determined by the encoders. 

The KL divergence items in Eq. \ref{eq:elbo_fact} have analytical form. 
As for the first two expectation terms in Eq.\ref{eq:elbo_fact}, we could apply  the reparameterization trick  and stochastic gradient variational Bayes (SGVB)  \cite{kingma2013auto} to compute the expectations with the Monte Carlo estimation. 
Hence, the ELBO of Eq. \ref{eq:elbo_fact} could be written as:

\begin{equation}\label{eq:elbo_detailed_form}
\begin{aligned}
\begin{array}{l}

\mathcal{L}_{e}(\boldsymbol{\theta},{\boldsymbol{{\phi}}};\mathbf{A},\mathbf{X}) \\
=
\dfrac{1}{LN^2}\sum\limits_{l=1}^{L}\sum\limits_{i,j\in \mathcal{V}}\left[\log p_{\boldsymbol{\theta}_1}(\mathbf{A}_{ij}\ |\  \mathbf{Z}^{\mathcal{V}(l)}_i,\mathbf{Z}^{\mathcal{V}(l)}_j) \right] \\
+\dfrac{1}{LMN}\sum\limits_{l=1}^{L}\sum\limits_{i\in \mathcal{A},j\in \mathcal{V}} \left[\log p_{\boldsymbol{\theta}_2}(\mathbf{X}_{ij}\ |\ \mathbf{Z}_i^{\mathcal{A}(l)},\mathbf{Z}_j^{\mathcal{V}(l)})\right] \\
-\dfrac{1}{2MJ}\sum\limits_{i=1}^{M}\sum\limits_{j=1}^{J}\left[\dfrac{\boldsymbol{\sigma}^2_{\mathcal{A}_{i}}|_j}{\hat{\boldsymbol{\sigma}}^2_{\mathcal{A}_{i}}|_j}+\dfrac{\boldsymbol{\mu}^2_{\mathcal{A}_{i}}|_j}{\hat{\boldsymbol{\sigma}}^2_{\mathcal{A}_{i}}|_j} - \log \dfrac{\boldsymbol{\sigma}^2_{\mathcal{A}_{i}}|_j}{\hat{\boldsymbol{\sigma}}^2_{\mathcal{A}_{i}}|_j}-1 \right ] \\
- \dfrac{1}{2NKJ} \sum\limits_{i=1}^{N} \sum\limits_{c=1}^{K}\gamma_{i,c}  \sum\limits_{j=1}^{J}\left[\dfrac{\boldsymbol{\sigma}^2_{\mathcal{V}_{i}}|_j}{\boldsymbol{\hat{\sigma}}^2_{c}|_j}+\dfrac{(\boldsymbol{\mu}_{\mathcal{V}_{i}}|_j-\boldsymbol{\hat{\mu}}_c|_j)^2}{\boldsymbol{\sigma}^2_{c}|_j} \right. \\ \left. -\log \dfrac{\boldsymbol{\sigma}^2_{\mathcal{V}_{i}}|_j}{\boldsymbol{\hat{\sigma}}^2_{c}|_j}-1\right] 
-  \dfrac{1}{NK}\sum\limits_{i=1}^{N}\sum\limits_{c=1}^{K}\gamma_{i,c} \log \dfrac{\gamma_{i,c}}{\pi_c},\\ \\ 

\text{where}\\ \\ 

\mathbf{Z}_{i}^{\mathcal{V}^{(l)}}=\boldsymbol{\mu}_{\mathcal{V}_{i}}+\boldsymbol{\sigma}_{\mathcal{V}_{i}} \odot \boldsymbol{\epsilon}^{(l)}, \text { with } \boldsymbol{\epsilon}^{(l)} \sim \mathcal{N}(0, \mathbf{I}) \\
\mathbf{Z}_{j}^{\mathcal{V}^{(l)}}=\boldsymbol{\mu}_{\mathcal{V}_{j}}+\boldsymbol{\sigma}_{\mathcal{V}_{j}} \odot \boldsymbol{\epsilon}^{(l)}, \text { with } \boldsymbol{\epsilon}^{(l)} \sim \mathcal{N}(0, \mathbf{I}) \\
\mathbf{Z}_{i}^{\mathcal{A}(l)}=\boldsymbol{\mu}_{\mathcal{A}_{i}}+\boldsymbol{\sigma}_{\mathcal{A}_{i}} \odot \boldsymbol{\epsilon}^{(l)}, \text { with } \boldsymbol{\epsilon}^{(l)} \sim \mathcal{N}(0, \mathbf{I}).

\end{array}
\end{aligned}
\end{equation}

In Eq. \ref{eq:elbo_detailed_form}, $L$ represents the size of Monte Carlo sampling to estimate the expectations and  $\boldsymbol{\cdot}|_j$ denotes the $j^{th}$ element of the vector. 
$\gamma_{i,c} $ represents $q_{{\boldsymbol{{\phi}}}_3}(\mathbf{C}_{i,c}|\mathbf{A}_i,\mathbf{X}_i)$ for simplicity. 
$\odot$ denotes the element-wise product between the auxiliary noise from   $\mathcal{N}(0,\mathbf{I})$ and the posterior distribution. 
To optimize Eq. \ref{eq:elbo_detailed_form}, we need to choose detailed inference network $f_{\boldsymbol{{\phi}}}$ and generative network $g_{\boldsymbol{\theta}}$ as encoders and decoders to simulate $q_{\boldsymbol{\phi}}$ and  $p_{\boldsymbol{\theta}}$, respectively.

\textbf{Inference network $f_{\boldsymbol{{\phi}}}$}. As shown in Fig. \ref{fig:framework}, the inference network $f_{\boldsymbol{{\phi}}}$ is parameterized by a two-layer GCN  and a two-layer multilayer perceptron (MLP) for inferring the means and variances of $q_{{\boldsymbol{{\phi}}}_1}(\mathbf{Z}^\mathcal{A}\ |\ \mathbf{X})$ and  $q_{{\boldsymbol{{\phi}}}_2}(\mathbf{Z}^\mathcal{V}\ |\ \mathbf{A},\mathbf{X})$, respectively.
Specifically,  the two-layer GCN inference network is organized as:

\begin{equation}\label{eq:gcn_encoder}
\begin{array}{l}
\mathbf{H}_{\mathcal{V}}^{(1)}=\operatorname{ReLU}\left(\tilde{\mathbf{A}} \mathbf{X} \mathbf{W}_{\mathcal{V}}^{(0)}\right) \\
{\left[\boldsymbol{\mu}_{\mathcal{V}}, \boldsymbol{\sigma}_{\mathcal{V}}^{2}\right]=\tilde{\mathbf{A}} \mathbf{H}_{\mathcal{V}}^{(1)} \mathbf{W}_{\mathcal{V}}^{(1)}},
\end{array}
\end{equation}

where $\left[\boldsymbol{\mu}_{\mathcal{V}}, \boldsymbol{\sigma}_{\mathcal{V}}^{2}\right]$ are the learned parameters for the posterior approximation distribution $q_{{\boldsymbol{{\phi}}}_2}(\mathbf{Z}^\mathcal{V}|\mathbf{A},\mathbf{X})$. 
$\text{ReLU}(\boldsymbol{\cdot})=max(0,\cdot)$ is rectified linear unit, a popular activation function.
${\boldsymbol{{\phi}}}_2$ = $[ \mathbf{W}_{\mathcal{V}}^{(0)}, \mathbf{W}_{\mathcal{V}}^{(1)}]$ are the trainable weight matrices for the two-layer GCN.
$\tilde{\mathbf{A}} = \mathbf{D}^{-\frac{1}{2}}\mathbf{A}\mathbf{D}^{-\frac{1}{2}}$  is the symmetrically normalized adjacency matrix of $\mathcal{G}$ with $\mathbf{D}_{ii}=\sum_j\mathbf{A}_{ij}$ being the degree matrix.

The two-layer MLP is defined as: 

\begin{equation}\label{eq:attribute_encoder}
\begin{array}{l}
\mathbf{H}_{\mathcal{A}}^{(1)}=\tanh \left(\mathbf{X}^{{T}} \mathbf{W}_{\mathcal{A}}^{(0)}+\mathbf{b}^{(0)}\right)  \\
{\left[\boldsymbol{\mu}_{\mathcal{A}}, \boldsymbol{\sigma}_{\mathcal{A}}^{2}\right]=\mathbf{H}_{\mathcal{A}}^{(1)} \mathbf{W}_{\mathcal{A}}^{(1)}+\mathbf{b}^{(1)}},
\end{array}
\end{equation}

where  $\left[\boldsymbol{\mu}_{\mathcal{A}}, \boldsymbol{\sigma}_{\mathcal{A}}^{2}\right]$ are the parameters for the posterior approximation distribution $q_{{\boldsymbol{{\phi}}}_1}(\mathbf{Z}_i^\mathcal{A}|\mathbf{X})$. tanh($\boldsymbol{\cdot}$) denotes the tangent activation function.
${\boldsymbol{{\phi}}}_1=[\mathbf{W}_{\mathcal{A}}^{(0)}, \mathbf{W}_{\mathcal{A}}^{(1)}, \mathbf{b}^{(0)}, \mathbf{b}^{(1)}]$ are the trainable weights in the two- layer  MLP encoder. 
Based on the Gaussian embeddings learned by the inference networks, the reparameterization trick \cite{kingma2013auto,kipf2016variational} will be used to reparameterize the latent embeddings for nodes and attributes in a differentiable way to jointly optimize the inference network and generative network. 

\textbf{Generative network $g_{\boldsymbol{\theta}}$}. 
The  generative network is used to simulate  $p_{\boldsymbol{\theta}_1}(\mathbf{A}\ |\  \mathbf{Z}^{\mathcal{V}})$ (\abbr $p_{\boldsymbol{\theta}_1}$) and $p_{\boldsymbol{\theta}_2}(\mathbf{X}\ |\ \mathbf{Z}^{\mathcal{A}}, \mathbf{Z}^{\mathcal{V}})$ (\abbr $p_{\boldsymbol{\theta}_2}$) 
to maximize  the likelihood of the observed variable $\mathbf{A}$ and $\mathbf{X}$.  
Since both the adjacent matrix $\mathbf{A}$ and the feature matrix $\mathbf{X}$ of the graph $\mathcal{G}$ are binary-valued in our experimental datasets, we set the generative network to learn the best parameters of Bernoulli distributions for the observed nodes and attributes as: 
\begin{equation}\label{eq:decoder}
    \begin{array}{l}
    p_{\boldsymbol{\theta}_{1}}\left(\mathbf{A}_{i j} \mid \mathbf{Z}_{i}^{\mathcal{V}}, \mathbf{Z}_{j}^{\mathcal{V}}\right)=\operatorname{Ber}\left(\boldsymbol{\tilde{\mu}}_{\mathcal{V}_{i j}}\right),\  \boldsymbol{\tilde{\mu}}_{\mathcal{V}_{i j}} = g_{\boldsymbol{\theta}_1}(\mathbf{Z}_i^\mathcal{V},\mathbf{Z}_j^\mathcal{V}) \\
    p_{\boldsymbol{\theta}_{2}}\left(\mathbf{X}_{i j} \mid \mathbf{Z}_{i}^{\mathcal{V}}, \mathbf{Z}_{j}^{\mathcal{A}}\right)=\operatorname{Ber}\left(\boldsymbol{\tilde{\mu}}_{\mathcal{A}_{i j}}\right),\  \boldsymbol{\tilde{\mu}}_{\mathcal{A}_{i j}} = g_{\boldsymbol{\theta}_2}(\mathbf{Z}_i^\mathcal{V},\mathbf{Z}_j^\mathcal{A}), \\ \\ 
    \text{where} \\  \\ 
    g_{\boldsymbol{\theta}_1}(\mathbf{Z}_i^\mathcal{V},\mathbf{Z}_j^\mathcal{V}) = \operatorname{sigmoid}(\mathbf{Z}_i^{\mathcal{V}T}\mathbf{Z}_j^\mathcal{V}) \\ 
    g_{\boldsymbol{\theta}_2}(\mathbf{Z}_i^\mathcal{V},\mathbf{Z}_j^\mathcal{A}) = \operatorname{sigmoid}(\mathbf{Z}_i^{\mathcal{V}T}\mathbf{Z}_j^\mathcal{A}). 
    \end{array}
\end{equation}

The encoder $g_{\boldsymbol{\theta}_1}(\boldsymbol{\cdot})$ and $g_{\boldsymbol{\theta}_2}(\boldsymbol{\cdot})$ use simple inner product between latent variables with sigmoid activation function. 
Hence, the $\log$ likelihood of $p_{\boldsymbol{\theta}_{1}}\left(\mathbf{A}_{i j} \mid \mathbf{Z}_{i}^{\mathcal{V}}, \mathbf{Z}_{j}^{\mathcal{V}}\right)$ and $p_{\boldsymbol{\theta}_{2}}\left(\mathbf{X}_{i j} \mid \mathbf{Z}_{i}^{\mathcal{V}}, \mathbf{Z}_{j}^{\mathcal{A}}\right)$ can be computed as follows: 
\begin{equation}\label{eq:bern_log_likelihood}
    \begin{array}{l}
        p_{\boldsymbol{\theta}_{1}} = \sum\limits_{d=1}^{J} \mathbf{A}_{ij}\log(\tilde{\boldsymbol{u}}_{\mathcal{V}_{ij}})+(1-\mathbf{X}_{ij})(1-\tilde{\boldsymbol{u}}_{\mathcal{V}_{ij}}) \\ 
         p_{\boldsymbol{\theta}_{2}} = \sum\limits_{d=1}^{J} \mathbf{X}_{ij}\log(\tilde{\boldsymbol{u}}_{\mathcal{A}_{ij}})+(1-\mathbf{X}_{ij})(1-\tilde{\boldsymbol{u}}_{\mathcal{A}_{ij}})  \\ 
    \end{array}
\end{equation}

For the real-value weighted datasets, we could replace the Bernoulli distribution with Gaussian distribution.




\subsection{Node Clustering Inference}

The latent variable $\mathbf{C}$ that represents the chosen Gaussian priors for latent node embeddings is used for node clustering assignments.  
To infer $\mathbf{C}$, we need to derive the analytical form for $q_{{\boldsymbol{{\phi}}}_3}(\mathbf{C}\ |\ \mathbf{A},\mathbf{X})$. 
However, directly inferring $\mathbf{C}$ is nontrivial since it is difficult to find an appropriate distribution for $q_{{\boldsymbol{{\phi}}}_3}(\mathbf{C}\ |\ \mathbf{A},\mathbf{X})$.
Therefore, similar to VaDE \cite{jiang2017variational}, we resort to find an approximation distribution for $q_{{\boldsymbol{{\phi}}}_3}(\mathbf{C}\ |\ \mathbf{A},\mathbf{X})$.

The evidence lower bound of $\log p(\mathbf{A},\mathbf{X})$ in Eq. \ref{eq:elbo} could  be rewritten as: 

\begin{equation}\label{eq:elbo_cluster_index}
    \begin{aligned}
    \begin{array}{l}
         \mathcal{L}_{e}(\boldsymbol{\theta},{\boldsymbol{{\phi}}};\mathbf{A},\mathbf{X}) \\ 

         = \mathbb{E}_{ q_{\boldsymbol{{\phi}}}} \left[\log \dfrac{p(\mathbf{A},\mathbf{X} \ |\  \mathbf{Z}^{\mathcal{A}},\mathbf{Z}^{\mathcal{V}},\mathbf{C}) p( \mathbf{Z}^{\mathcal{A}},\mathbf{Z}^{\mathcal{V}},\mathbf{C})}{q_{\boldsymbol{{\phi}}}(\mathbf{Z}^{\mathcal{A}},\mathbf{Z}^{\mathcal{V}},\mathbf{C}\ |\ \mathbf{A},\mathbf{X})}\right] \\ 
         = \mathbb{E}_{ q_{\boldsymbol{{\phi}}}} \left[\log \dfrac{p(\mathbf{A},\mathbf{X} \ |\  \mathbf{Z}^{\mathcal{A}},\mathbf{Z}^{\mathcal{V}},\mathbf{C}) p( \mathbf{Z}^{\mathcal{A}})p(\mathbf{Z}^{\mathcal{V}})p(\mathbf{C} \ |\ \mathbf{Z}^{\mathcal{V}})}{q_{\boldsymbol{{\phi}}}(\mathbf{Z}^{\mathcal{A}},\mathbf{Z}^{\mathcal{V}},\mathbf{C}\ |\ \mathbf{A},\mathbf{X})}\right] \\     
         = \mathbb{E}_{ q_{\boldsymbol{{\phi}}}} \left[\log \dfrac{p(\mathbf{A},\mathbf{X} \ |\  \mathbf{Z}^{\mathcal{A}},\mathbf{Z}^{\mathcal{V}},\mathbf{C}) p( \mathbf{Z}^{\mathcal{A}})p(\mathbf{Z}^{\mathcal{V}})}{q_{\boldsymbol{{\phi}}}(\mathbf{Z}^{\mathcal{A}},\mathbf{Z}^{\mathcal{V}} \ |\ \mathbf{A},\mathbf{X})}\right] \\ 
         -\mathbb{E}_{q_{{\boldsymbol{{\phi}}}}} \bigg[ D_{KL}(q_{{\boldsymbol{{\phi}}}_3}(\mathbf{C}\ |\ \mathbf{A},\mathbf{X})\ \|  \ p(\mathbf{C} \ |\ \mathbf{Z}^{\mathcal{V}}) \bigg] \\
        
    \end{array}
    \end{aligned}
\end{equation}

In Eq. \ref{eq:elbo_cluster_index}, the expectation of KL divergence \wrt $q_{\boldsymbol{{\phi}}}$  is non-negative and it becomes zero only if the two distributions in  $D_{KL}$ are identical. 
Therefore, the \textit{ELBO} of $\log p(\mathbf{A},\mathbf{X})$ could become even tighter if $ q_{{\boldsymbol{{\phi}}}_3}(\mathbf{C}\ \|  \mathbf{A},\mathbf{X}) = p(\mathbf{C} \ |\ \mathbf{Z}^{\mathcal{V}})$. 
We assume $q_{{\boldsymbol{{\phi}}}_3}(\mathbf{C}\ |\ \mathbf{A},\mathbf{X}) = \prod\limits_{i=1}^{N}q_{{\boldsymbol{{\phi}}}_3}(\mathbf{C}_i\ |\ \mathbf{A}_i,\mathbf{X}_i) $ and  $p(\mathbf{C}\ |\ \mathbf{Z}^{\mathcal{V}})=\prod\limits_{i=1}^{N} p(\mathbf{C}_i\ |\ \mathbf{Z}^{\mathcal{V}}_i)$.
Thus,  we set $q_{{\boldsymbol{{\phi}}}_3}(\mathbf{C}_i\ |\ \mathbf{A}_i,\mathbf{X}_i) =  p(\mathbf{C}_i\ |\ \mathbf{Z}^{\mathcal{V}}_i) $ and get  $q_{{\boldsymbol{{\phi}}}_3}(\mathbf{C}\ |\ \mathbf{A},\mathbf{X})= p(\mathbf{C}\ |\ \mathbf{Z}^{\mathcal{V}})$. 
According to the Bayes' rule, the probability for $  p(\mathbf{C}_i\ |\ \mathbf{Z}^{\mathcal{V}}_i)$ could be computed as:
\begin{equation} 
    \label{eq:prior_choosing}
  p(\mathbf{C}_i\ |\ \mathbf{Z}^{\mathcal{V}}_i) = \dfrac{p(\mathbf{Z}_i^{\mathcal{V}}\ |\ \mathbf{C}_i)p(\mathbf{C}_i)}{\sum\limits_{\mathbf{C}}p(\mathbf{Z}_i^{\mathcal{V}}\ |\ \mathbf{C})p(\mathbf{C})},
\end{equation}

where $p(\mathbf{Z}_i^{\mathcal{V}}\ |\ \mathbf{C}_i)p(\mathbf{C}_i)$ is based on the generative process for nodes with  Gaussian mixture model as priors. 
It is worth noting that the calculation for $p(\mathbf{C}_i\ |\ \mathbf{Z}^{\mathcal{V}}_i)$ is in the embeddings space rather than the raw features of the nodes.
Hence, it becomes clear that the clustering assignments of nodes could be inferred simultaneously with the variational co-embedding learning process.




\subsection{Clustering Assignment Hardening}


In order to further strengthen the clustering  performance using the Gaussian mixture priors, we exploit a clustering assignment hardening loss from DEC \cite{xie2016unsupervised} to put more emphasis  on the highly confident node clustering assignment.

First, we compute a soft assignment distribution $\mathbf{Q}$ for each node against different Gaussian prior centers using Eq. \ref{eq:soft_assignment_q}.
The node  probability  distribution against  each Gaussian prior is measured by Student's $t$-distribution. $\alpha$ is the degree of freedom and is set as one in our experiments. 

\begin{equation}\label{eq:soft_assignment_q}
    \mathbf{Q}_{i c}=\dfrac{\left(1+\left\|\mathbf{Z}_{i}^{\mathcal{V}}-\boldsymbol{\mu}_{c}\right\|^{2} / \alpha\right)^{-\frac{\alpha+1}{2}}}{\sum_{c^{\prime}}\left(1+\left\|\mathbf{Z}_{i}^{\mathcal{V}}-\boldsymbol{\mu}_{c^{\prime}}\right\|^{2} / \alpha\right)^{-\frac{\alpha+1}{2}}} 
\end{equation}

\begin{equation}\label{eq:soft_assignment_p}
    \mathbf{P}_{i c}=\dfrac{\mathbf{Q}_{i c}^{2} / \Sigma_{i} \mathbf{Q}_{i c}}{\sum_{c^\prime}\mathbf{Q}_{i c^{\prime}}{}^{2} / \Sigma_{i} \mathbf{Q}_{i c^{\prime}}}
\end{equation}

Then, the soft assignment distribution $\mathbf{Q}$ is further squared and normalized to have stricter probabilities with Eq. \ref{eq:soft_assignment_p}. 
Finally, the KL divergence is adopted to between $\mathbf{P} $ and $\mathbf{Q}$ using Eq. \ref{eq:soft_assignment_kl} to push  $\mathbf{Q}$ put more emphasis on the highly confident soft assignment by optimizing the neural network parameters and the trainable Gaussian mixture priors. 

\begin{equation}\label{eq:soft_assignment_kl}
    \mathcal{L}_h=\mathrm{KL}(\mathbf{P} \| \mathbf{Q})=\sum_{i} \sum_{c} \mathbf{P}_{i c} \log \frac{\mathbf{P}_{i c}}{\mathbf{Q}_{i c}}
\end{equation}

\subsection{Mutual Distance for Gaussian Priors}

To avoid the Gaussian mixture priors collapsing to each other during the training process, we further design a mutual distance loss to force the Gaussian centers of different priors to move away from  each other.
As a result, the regularized latent node embedding distributions by Gaussian mixture priors could become more distinguishable for different node categories.

\begin{equation}\label{eq:mutual_distance}
    \mathcal{L}_{m}=\dfrac{1}{K^2}\sum\limits_{c=1}^{K} \sum\limits_{k=1}^{K} \|\boldsymbol{\mu}_{c}-\boldsymbol{\mu}_{k}\|_2
\end{equation}

The mutual distance between Gaussian priors is defined in Eq. \ref{eq:mutual_distance}, which averaging the Euclidean distance among all the Gaussian  centers. 
During the training process, the mutual distance loss is maximized through updating the trainable means of different Gaussian priors.

\subsection{Model Training}

Our final objective function unifies the \textit{ELBO} $\mathcal{L}_e$, the cluster assignment hardening loss  $\mathcal{L}_h$ and the mutual distance loss $\mathcal{L}_m$
to jointly optimize the parameters of the proposed model.

\begin{equation} \label{eq: objective_function} 
    \mathcal{L}=\mathcal{L}_e +\beta \mathcal{L}_m - \omega \mathcal{L}_h
\end{equation}

As shown in Eq. \ref{eq: objective_function}, $\omega$ and $\beta$ are the weight coefficients to adjust the contributions of the cluster assignment hardening loss (Eq. \ref{eq:soft_assignment_kl}) and the mutual distance loss (Eq. \ref{eq:mutual_distance}).
We optimize Eq. \ref{eq: objective_function} with gradient descent based optimizer to update the model parameters \wrt $\{{\boldsymbol{{\phi}}},\boldsymbol{\theta}, \boldsymbol{\mu}, \boldsymbol{\sigma}, \boldsymbol{\pi}\}$. 
${\boldsymbol{{\phi}}}$ and $\boldsymbol{\theta}$ are the trainable weights for the inference and generative networks. $\{\boldsymbol{\mu}, \boldsymbol{\sigma}, \boldsymbol{\pi}\}$ are the trainable parameters of the Gaussian mixture priors.

We first  train the inference and generative network for node embeddings without updating the Gaussian parameters. 
Then, an external GMM cluster model is applied on the node embeddings to initialize the parameters of the internal Gaussian  priors in VCLANC.
Finally, we optimize the network parameters and  the Gaussian parameters intermittently.
The final clustering assignment for each node could be derived based on Eq. \ref{eq:prior_choosing}.
Algorithm \ref{algorithm} summarizes our proposed method.
From line 1 to line 3, the algorithm only updates the parameters $\{\boldsymbol{{\phi}}$, $\boldsymbol{\theta}\}$ for the inference and generative network  to get initial node embeddings.
Then, the internal Gaussian prior parameters $\{\boldsymbol{\mu}, \boldsymbol{\sigma},\boldsymbol{\pi}\}$ are initialized by GMM from the current node embedding space in line 4.
After that, the algorithm performs iterative updating for the neural networks and the Gaussian mixture priors to form better clusters. Finally, the node embeddings  $\mathbf{Z}^{\mathcal{V}}$ and the clustering assignments could be easily derived based on the trained network and the Gaussian mixture priors.




\begin{algorithm}[!t]
\SetAlgoLined
\KwIn{attributed graph $\mathcal{G}$; embedding size $J$; hyper-parameters $\omega$ and $\beta$; training epochs $T_1$ and $T_2$; Gaussian  parameter updating interval $T$.
}

\KwOut{node embedding $\mathbf{Z^{\mathcal{V}}}$; node clustering assignments.}

\For{$epoch \leftarrow$ 1 \KwTo $T_1$ }
{
update $\{\boldsymbol{{\phi}}$, $\boldsymbol{\theta}\}$. 
}
Apply GMM on $\mathbf{Z}^{\mathcal{V}}$ to get $\{\boldsymbol{\mu},  \boldsymbol{\sigma},\boldsymbol{\pi}\}$.

\For{$epoch \leftarrow$ 1 \KwTo $T_2$}
{
\If{$epoch\%10 <  T$}
{
update $\{\boldsymbol{{\phi}}$, $\boldsymbol{\theta}\}$.
}
\Else{ 
update $\{\boldsymbol{\mu}, \boldsymbol{\sigma},\boldsymbol{\pi}\}$.}
}
Compute clustering assignment using Eq. \ref{eq:prior_choosing}. 
\caption{VCLANC}
\label{algorithm}
\end{algorithm}

\section{Experimental Setup}

In this section, we introduce our experimental setup  including the compared baselines and parameter settings, the adopted datasets and evaluation metrics.



\subsection{Baselines and Parameter Settings}


We introduce all the adopted baselines and their parameter settings as follows:

\begin{itemize}
	\item $K$-means \cite{macqueen1967some} is a partition based clustering algorithm to iteratively divide the data into $K$ number of groups and  update the cluster centers until the cluster centers become stable.
	\item Gaussian mixture model (GMM) is a probabilistic clustering model 
	which assumes the data  are generated from a  Gaussian mixture distributions and the cluster indexes for the data  are determined by the component assignment probability.
	\item GAE \cite{kipf2016variational} is an unsupervised embedding learning method for graph-structured data based on the  auto-encoder  framework.
	GAE adopts GCN as encoder and an inner product as decoder. 
	\item VGAE \cite{kipf2016variational} is a probabilistic variant of GAE. 
	Compared with GAE, VGAE  assumes a prior distribution on the latent node embeddings and  optimizes the model through minimizing the KL divergence regularized  reconstruction loss between inputs and outputs.
	\item CAN \cite{Meng2019} is an extension of VGAE. 
	The difference between VGAE and CAN is that CAN jointly learns the embeddings for nodes and attributes using an extra auto-encoder network to process the node attributes.
    \item SDCN \cite{bo2020www} uses a delivery operator to transfer the knowledge learned by a deep autoencoder to the corresponding deep GCN. Then, a dual self-supervised mechanism is used to unify these two modules and guide the update of the whole network.
    \item NEC \cite{sun2020network} is  an extension of GAE to detect the attributed network communities  using a soft clustering assignment loss and a network modularity loss.
\end{itemize}

The parameter settings for all the methods are detailed as follows.  
The main parameter for $K$-means and GMM is the number of clusters, which is set as the actual cluster size of the datasets. The other parameters of $K$-means and GMM are set as default following scikit-learn package \footnote{https://scikit-learn.org/stable/index.html}.
We adopt the same parameter settings for GAE, VGAE, CAN, NEC and VCLANC on all the datasets. 
Specifically, the total training epoch is set as 300 and   Adam optimizer \cite{kingma2014adam} is adopted with the learning rate as 0.002.
The two-layer GCN is set with the hidden dimension as 64 and 32, respectively.
For SDCN, we follow their parameter settings with the dimension of the latent layers set as 500-500-2000-10, the learning rate set as 0.0001 and epochs set as 50 \cite{bo2020www}. 
For the proposed method, we first train the model for 200 epochs only using the ELBO $\mathcal{L}_e$. Then, we use GMM to initialize the parameters for the Gaussian mixture priors.
After that, we train the model with the complete objective function in Eq. \ref{eq: objective_function} and  update the Gaussian parameters intermittently with $T=5$ in the remaining 100 epochs (shown in Algorithm \ref{algorithm}).
The two hyper-parameters  $\omega$ and $\beta$ are set as one. 
We use the machine learning library Pytorch \footnote{https://pytorch.org/} to implement the neural network related methods.


\subsection{Datasets}

\begin{table}[t]
    \centering
    \caption{Statistics of the datasets.}
    \begin{tabular}{|c|c|c|c|c|} \hline 
        Dataset & \#Nodes &\#Edges &\#Attributes & \#Clusters   \\ \hline 
        Cora & 2,708 & 5,429 & 1,433 & 7 \\ \hline
        Citeseer & 3,327 & 4,732 & 3,703 & 6 \\ \hline
        BlogCatalog & 5,196 & 171,743 & 8,189 & 6 \\ \hline
        Flickr & 7,575 & 239,738 & 12,047 & 9 \\ \hline
    \end{tabular}
    \label{tab:datasets}
\end{table}

We conduct experiments on four attributed network datasets with the  cluster size of nodes no less than six.
The statistics of the datasets are shown in Table \ref{tab:datasets}. 
The \textit{Cora} and {Citeseer} datasets are created from citation networks by Sen \etal \cite{sen2008collective}.  The nodes represents articles and the edges are their citation actions. Attributes of each node are a list keywords. 
The \textit{BlogCatalog} \cite{Blogcatalog_Tang_2009} and \textit{Flickr} \cite{Flickr_Huang_2017}  datasets are created from social networks. The edges represent the users' relationships and the attributes are the keywords extracted from the users' profiles. 
The ground-truth cluster information of nodes in each dataset is included and could be used to evaluate the clustering performance of the different methods.

\subsection{Evaluation Metrics}

We use six  metrics to  evaluate the clustering performance.
Suppose that the predicted partition and the true partition of a network are represented as 
$\Omega = \{\omega_1, \cdots, \omega_q\}$ and $\mathcal{C} = \{c_1, \cdots, c_p\}$, respectively. 
Larger score implies better clustering performance across all the metrics.
Each of the metrics are briefly introduced in the follow. 
\begin{itemize}
    \item \textbf{NMI} is metric used to determine the clustering quality by computing normalized mutual information between the predicted partitions and the true partitions. 
    Its definition is shown as follows: 
    \begin{equation}
\text{NMI}(\Omega, \mathcal{C}) = \dfrac{ \sum_{i,j}\dfrac{|w_i \cap c_j|}{N} \log \dfrac{N |w_i \cap c_j|}{|w_i||c_j|}}{
(- \sum_{i} \dfrac{|w_j|}{N}\log \dfrac{|w_i|}{N} -  \sum_{j} \dfrac{|c_j|}{N} \log \dfrac{|c_j|}{N})/2
}. 
\end{equation}

    \item \textbf{Purity}. Ideally, the predicted partition should contain only the data objects from one categories. This metric is used to evaluate the average pure level among all the predicted partitions. It is defined as follows:
    
    \begin{equation} 
    \text { Purity }(\Omega, \mathcal{C})=\dfrac{1}{N} \sum_{i=1}^{k} \max _{j}\left|c_{i} \cap \omega_{j}\right|.
    \end{equation}
    
    \item \textbf{ARI} is  an adjust version of Rand
Index which is used to calculate the percentage of the correct clustering assignments. ARI is defined as follows: 
\begin{equation}
\begin{split}
	&ARI(\Omega, \mathcal{C}) = \\
	&\dfrac{\sum_{i,j}\binom{|\omega_i \cap c_j|}{2}-[\sum_i\binom{|\omega_i|}{2}\sum_j\binom{|c_j|}{2}]/\binom{n}{2}}{\dfrac{1}{2}[\sum_i\binom{|\omega_i|}{2} + \sum_j\binom{|c_j|}{2}] - [\sum_i \binom{|\omega_i|}{2}\sum_j \binom{|c_j|}{2}]/\binom{n}{2}}, 
\end{split}
\end{equation}

    \item \textbf{Precision (P)} denotes the percentage of predicted results which are truly relevant for a class. 
    For a dataset with multiple classes, we use weighted precision score to avoid the  imbalanced issue.
    Let $TP_i$ denotes the true positive prediction number, $FP_i$ denotes the false positive prediction number, and $FN_i$ denotes the false negative prediction number for the $i^{th}$ class, respectively.
    The weighted precision is defined as:
    
    $$
    \text{Precision}=\sum_{i} \dfrac{|c_i|}{N}\dfrac{TP_i}{TP_i+FP_i}.
    $$
    
    \item \textbf{Recall (R)}  denotes the rate of relevant results that were truly identified for a class. For multi-class dataset, we use the weighted recall defined as: 
    
    $$
    \text{Recall}=\sum_{i} \dfrac{|c_i|}{N} \dfrac{TP_i}{TP_i+FN_i}.
    $$
    
    \item \textbf{F1-score (F1)} is the harmonic mean of precision and recall for a class. For multi-class dataset, the weighted F1-score is defined as: 
    
    $$
    \text{F1-score}= \sum_{i} \dfrac{|c_i|}{N}  \dfrac{2TP_i}{2TP_i+FP_i+FN_i}.
    $$
    
\end{itemize}

\section{Results and Analysis}
In this section, we report the experimental results for all the compared methods. 

\subsection{Clustering Performance}

Tables \ref{tab:cora}, \ref{tab:citeseer}, \ref{tab:BlogCatalog} and \ref{tab:Flickr} show the clustering performance of all the methods on the adopted datasets.
We repeat the experiment on each dataset for 10 times to report the average performance.
The best and  second best results are marked with \textit{boldface} and \textit{underline}, respectively.

As we can see from Table \ref{tab:cora} to Table \ref{tab:Flickr}, the proposed VCLANC consistently achieves the best or second  best clustering performance over its baselines on the adopted datasets.
Let us first look through the results from the first group of baselines, which include $K$-means and GMM methods.   
These two general clustering methods show quite poor 
performance across all the datasets.
For example, the NMI value of $K$-Means and GMM on Cora dataset is 0.143 and 0.0772, which is far behind VCLANC with its NMI as 0.5334.
Similarly, the F1-score of $K$-means and GMM is 0.3318 and 0.2226, which is also worse compared with VCLANC with F1-score as 0.7004 and CAN with its F1-score as 0.6769.
Interestingly, the precision  of $K$-means and GMM achieve surprised high score on Flickr dataset in Table \ref{tab:Flickr}. 
However, this do not denote that $K$-means and GMM are better than the other methods on Flickr dataset as their recall value is quite low.
$K$-means and GMM can only perform black-box style clustering on the input data and they are not flexible to fuse different types of information such as the network structure and the node attribute. 
But it has been proved that fusing both the network structure and the node attributes is useful for the attributed network clustering task.

\begin{table}[t]

    \centering
    \caption{Clustering performance on Cora dataset. }
    \begin{tabular}{c|c|c|c|c|c|c|c}
    \toprule
         & NMI &Purity & ARI & F1 & P & R  \\ \toprule 
         $K$-means&0.1430 &0.3811 &0.0734 &0.3318 &0.4257 &0.3242   \\ 
         GMM & 0.0772 &0.3310 &0.0294 &0.2266 &0.4026 &0.3133  \\ \midrule
         GAE &0.4597 &0.6526 &0.3538 &0.6291 &0.7272 &0.6058  \\ 
         VGAE &0.461 &0.6536 &0.3693 &0.6433 &0.72 &0.6175  \\ 
         CAN &\underline{0.4893} &\underline{0.6842} &\underline{0.4361} &\underline{0.6769} &\underline{0.7286} &\underline{0.6579} \\ \midrule
         SDCN & 0.2199 & 0.4944 & 0.1658 & 0.4397 & 0.5103 & 0.4111 \\ 
         NEC&0.4481 &0.6224 &0.3799 &0.5865 &0.6418 &0.5694  \\ 
         VCLANC & \textbf{0.5334} &\textbf{0.7085} &\textbf{0.4841} &\textbf{0.7004} &\textbf{0.7504} &\textbf{0.6816}  \\ 

    \bottomrule
    \end{tabular}
    \label{tab:cora}
\end{table}

\begin{table}[!t]
    \centering
     \caption{Clustering performance on Citeseer dataset. }
    \begin{tabular}{c|c|c|c|c|c|c|c}
    \toprule
          & NMI &Purity & ARI & F1 & P & R  \\ \toprule
         $K$-means& 0.2215 &0.4724 &0.1758 &0.4625 &0.5243 &0.4519    \\ 
         GMM &0.1756 &0.4036 &0.1287 &0.3444 &0.4687 &0.3949  \\ \midrule
         GAE &0.1885 &0.4235 &0.0799 &0.3834 &0.5579 &0.3808  \\ 
         VGAE & 0.2426 &0.5053 &0.1554 &0.4706 &0.58 &0.4618  \\  
         CAN & 0.3651 &0.6464 &0.3384 &0.5976 &0.6381 &0.5950  \\ \midrule
         SDCN & \underline{0.3842} & \underline{0.6631} & \underline{0.4022} & \underline{0.6460} & \underline{0.6454} & \underline{0.6567} \\ 
         NEC & 0.319 &0.5882 &0.2867 &0.5352 &0.5772 &0.5307 \\ 
         VCLANC  & \textbf{0.4007} &\textbf{0.6769} &\textbf{0.4070} &\textbf{0.6560} &\textbf{0.6791} &\textbf{0.6439}   \\ 

    \bottomrule
    \end{tabular}
    \label{tab:citeseer}
\end{table}

\begin{table}[t]
    \centering
     \caption{ Clustering performance on BlogCatalog dataset.}
    \begin{tabular}{c|c|c|c|c|c|c|c}
    \toprule
          & NMI &Purity & ARI & F1 & P & R  \\ \toprule
         $K$-means & 0.1659 &0.3249 &0.0679 &0.2632 &0.4481 &0.3079 \\ 
         GMM &0.1733 &0.3096 &0.0602 &0.2194 &0.459 &0.3042  \\ \midrule
         GAE &0.1389 &0.3449 &0.0866 &0.3009 &0.3191 &0.3104 \\  
         VGAE & 0.2609 &0.4622 &0.1721 &0.4278 &0.4846 &0.4289  \\ 
          CAN &\underline{0.2919} &\underline{0.505} &\underline{0.1996} &\underline{0.4556} &\underline{0.5101} &\underline{0.4600} \\ \midrule
         SDCN &  0.2743 & 0.4390 & 0.1947 & 0.3755 & 0.3790 & 0.4024 \\ 
         NEC &0.2735 &0.4929 &0.1977 &0.4347 &0.4533 &0.4355  \\ 
         VCLANC & \textbf{0.3036} &\textbf{0.5254} &\textbf{0.2455} &\textbf{0.4884} &\textbf{0.5146} &\textbf{0.4935}\\
    \bottomrule
    \end{tabular}
    \label{tab:BlogCatalog}
\end{table}

\begin{table}[t]
    \centering 
     \caption{Clustering performance on Flickr dataset.}
    \begin{tabular}{c|c|c|c|c|c|c|c}
    \toprule
          & NMI &Purity & ARI & F1 & P & R  \\ \toprule
         $K$-means &0.0552 &0.1313 &0.0006 &0.0484 &\underline{0.4459} &0.1294  \\ 
         GMM &0.048 &0.1282 &0.0003 &0.0444 &\textbf{0.4617} &0.1267  \\  \midrule           
         GAE &0.1521 &0.2828 &0.0783 &0.2322 &0.2693 &0.2529  \\  
         VGAE &0.1116 &0.2485 &0.0601 &0.2034 &0.2092 &0.2290  \\ 
         CAN &0.1763 &0.3083 &0.0990 &0.2736 &0.3018 &0.2936  \\ \midrule
         SDCN &  \textbf{0.2682} & \textbf{0.3154} & \textbf{0.1665} & 0.2488 & 0.3163 & \textbf{0.3125}\\ 
         NEC&0.1299 &0.2657 &0.0744 &0.2347 &0.2366 &0.2362   \\ 
         VCLANC & \underline{0.1850} &\underline{0.3127} &\underline{0.1216} &\textbf{0.2943} &0.3343 &\underline{0.3086} \\
    \bottomrule
    \end{tabular}
    \label{tab:Flickr}
\end{table}

The second group of baselines in our experiments include GAE, VGAE and CAN, which first learn node embeddings and then use $K$-means for the clustering task. 
As shown in Tables from \ref{tab:cora} to \ref{tab:Flickr}, 
the overall clustering performance of GAE, VGAE and CAN is better than directly using $K$-means or GMM.
The performance improvements show that learning low-dimensional representation for the network is quite important in network clustering.
For instance, the NMI value of GAE and VGAE is around 0.46 on Cora dataset which is much better than $K$-means and GMM. Similarly, the F1-score of GAE and VGAE is around 0.64 on Cora dataset which is also far better than $K$-means or GMM. 
Among GAE, VGAE and CAN, we can see that the performance of  GAE is slightly worse than VGAE and CAN.
CAN achieves better performance than VGAE and also ranks as the second best among all the methods on the Cora, Citeseer and 
BlogCatalog.   
Compared with VGAE, CAN adds extra module to embed the attributes of the network into the same semantic space to learn the affinities between attributes and network structures. 
The experimental results of CAN prove that its co-embedding strategy can learn better representations for the attributed network clustering task.

The third group of baselines includes SDCN and NEC. Together with our proposed VCLANC, these methods aim to jointly learn the network embeddings and infer the clustering assignments for the attributed networks.
As shown in Table \ref{tab:cora} to Table \ref{tab:BlogCatalog}, VCLANC outperforms SDCN and NEC on Cora, Citeseer and 
BlogCatalog.
Take BlogCatalog dataset as an example, due to its increased network size and attributes number, the clustering performance of all the methods on BlogCatalog dataset decrease rapidly.
Nevertheless, the NMI value and F1-score of the proposed  method are 0.3036 and 0.4884, which still rank the best among all the methods.
NEC and SDCN achieve similar results on NMI, ARI on BlogCatalog with the scores are around 0.27 and 0.19, respectively. 
But SDCN is much better than NEC and VCLANC on Flickr dataset in NMI, purity and ARI with the values as 0.2682, 0.3154 and 0.1665, respectively.
The advantages of SDCN lies in its combined DNN and GCN structure, in which the DNN learns the overall information and GCN learns the local neighborhood structure. But it may be inconvenient as the DNN need to be pre-trained. 
At the same time, SDCN is more complicated than other methods as it requires much bigger neuron number in each layer ( 500-500-2000-10), while VCLANC and NEC only use a two-layer GCN (64-32).

Compared with SDCN, NEC uses the standard auto-encoder with soft clustering assignment loss and NEC also adopts extra modularity loss to exploit the higher-order proximity of nodes into the representation learning.
However, NEC can not set different Gaussian priors to guide the embedding learning.
On the contrary, VCLANC assumes a Gaussian mixture priors to guide the embedding learning of different node categories.
Moreover, VCLANC also co-embeds the network and attribute into same semantic space to exploit the mutual affinities between nodes and attributes.
Hence, VCLANC learns better node embeddings and outperforms NEC. 
What is more, when comparing CAN with VCLANC, the latter adopts GMM as priors for joint clustering purpose, which leads to better clustering performance.
For example, the NMI of VCLANC is 0.5334 on Cora dataset while the NMI of CAN is 0.4893. 
As for the Citeseer dataset, VCLANC also show significant improvement over CAN with its ARI value as 0.407, which is higher than CAN with ARI as 0.3384.
Therefore, it proves that joint learning the node embeddings and clustering assignments is useful when comparing the results of CAN and VCLANC.

\subsection{Node Embedding Visualization}

\begin{figure}[!]
    \centering
    \subfloat[Raw attributes\label{<cora_kmeans_tsne>}]{\includegraphics[width=0.1\textwidth]{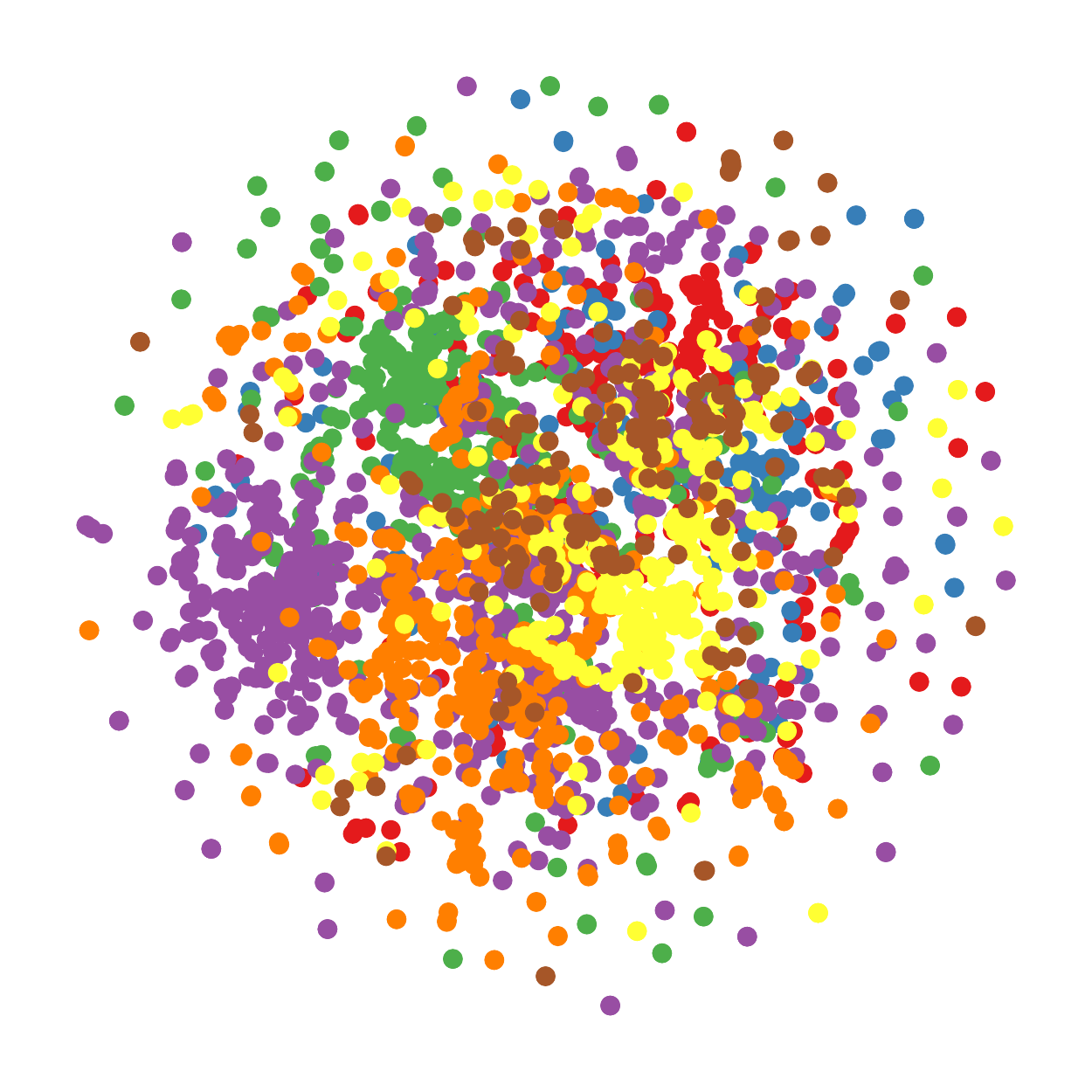}}\hfill
    \subfloat[GAE\label{<cora_gae_tsne>}]{\includegraphics[width=0.1\textwidth]{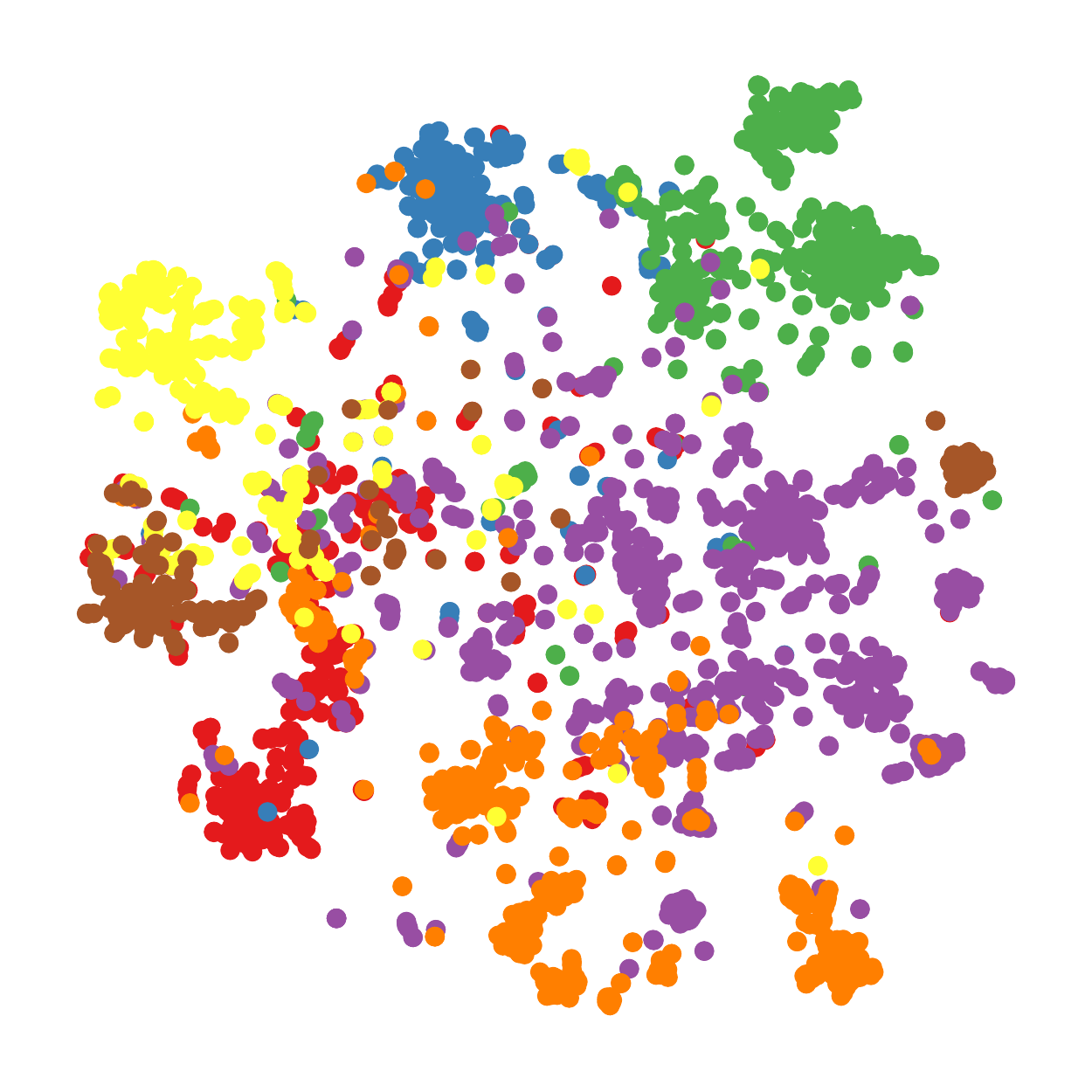}}\hfill 
    \subfloat[VGAE\label{<cora_vgae_tsne>}]{\includegraphics[width=0.1\textwidth]{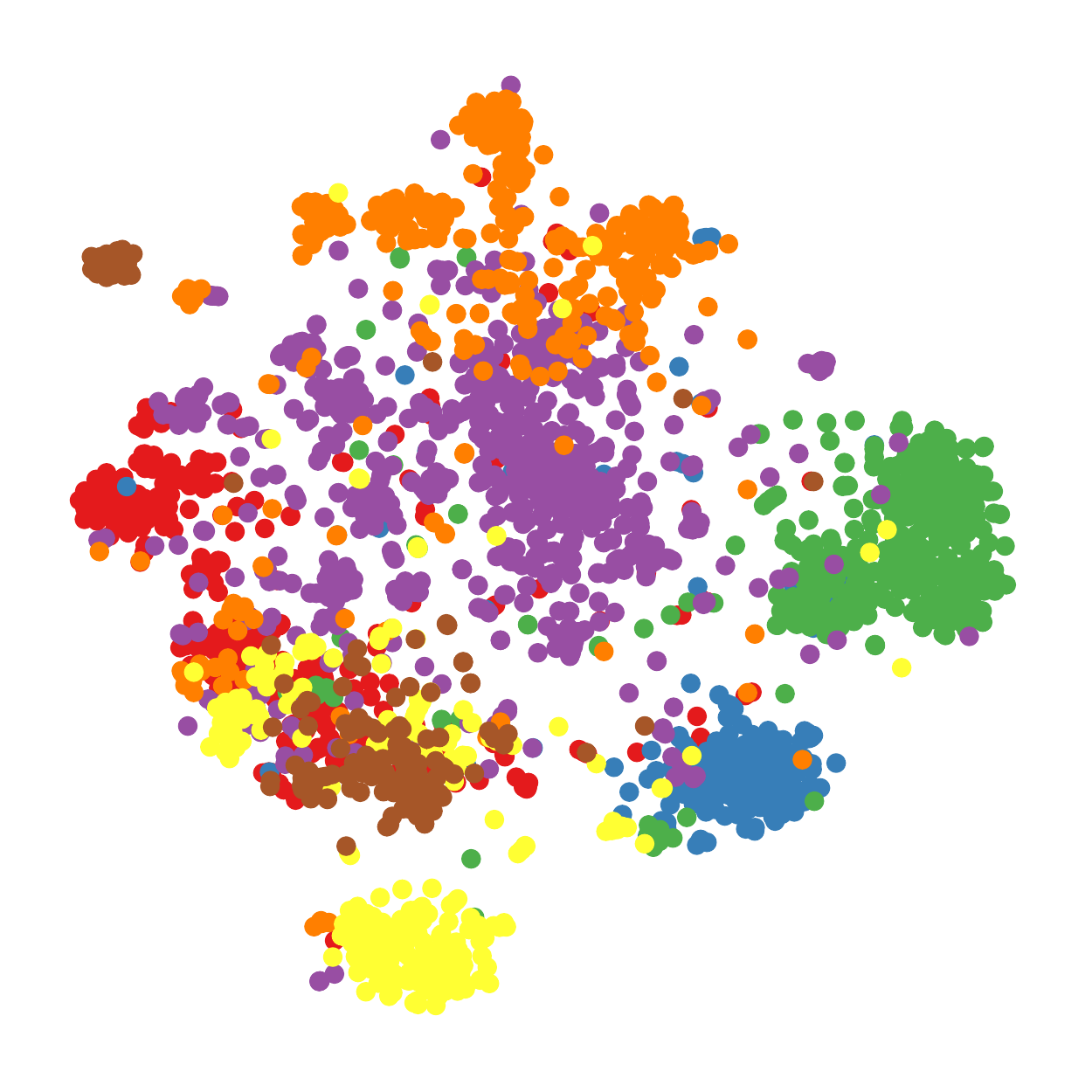}}\hfill
    \subfloat[CAN\label{<cora_gmm_tsne>}]{\includegraphics[width=0.1\textwidth]{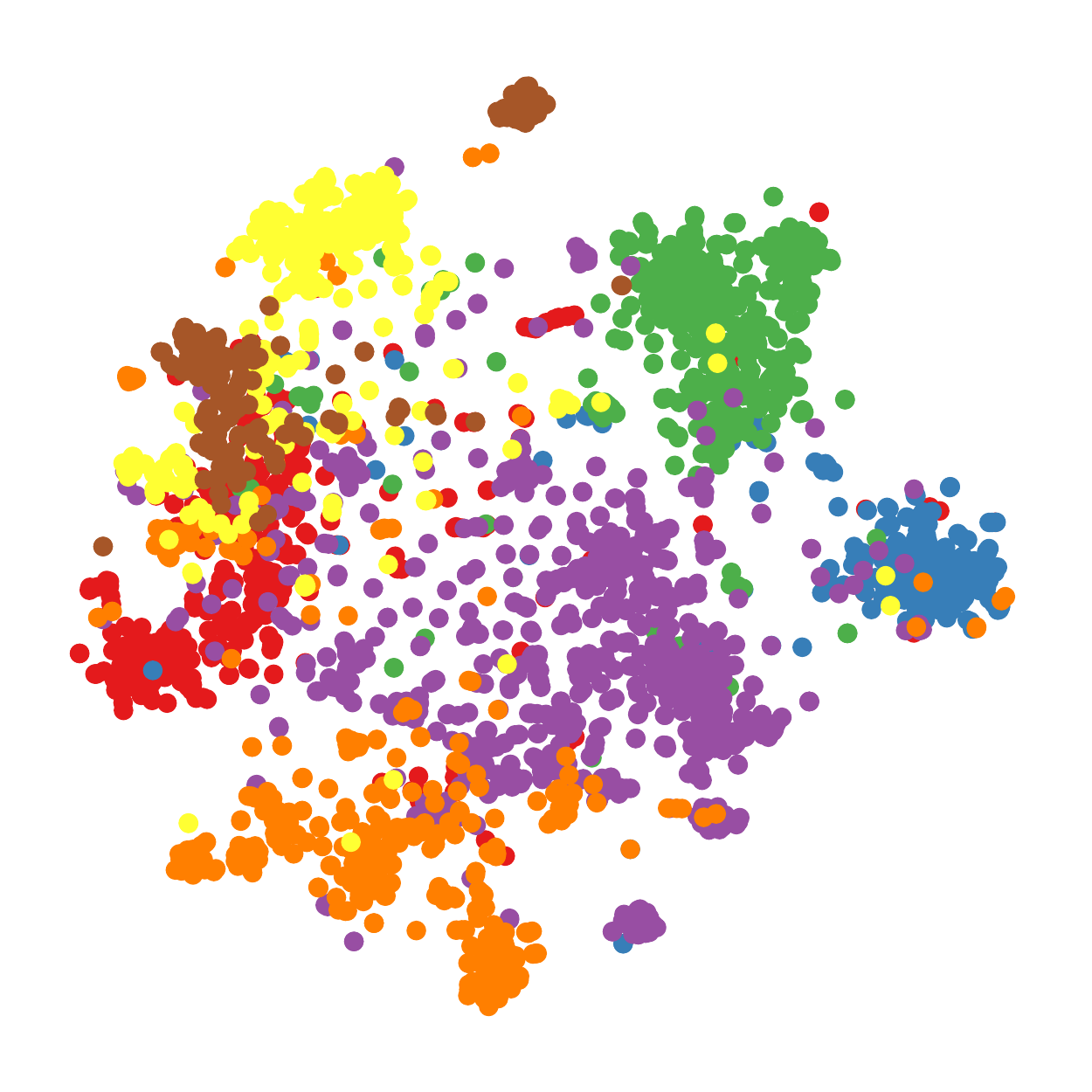}}\hfill \\
    \subfloat[SDCN\label{<cora_daegc_tsne>}]{\includegraphics[width=0.1\textwidth]{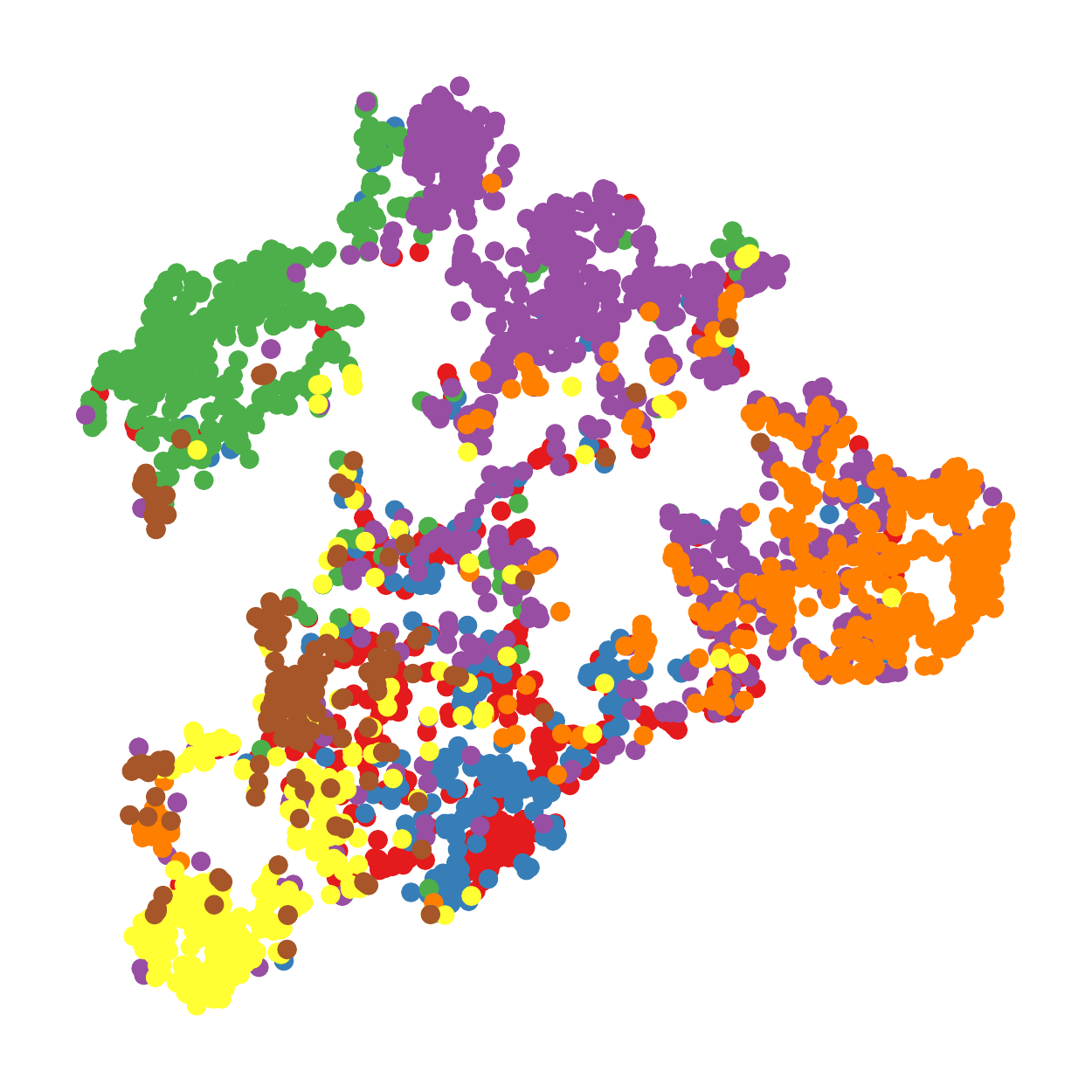}}\hfill
    \subfloat[NEC\label{<cora_nec_tsne>}]{\includegraphics[width=0.1\textwidth]{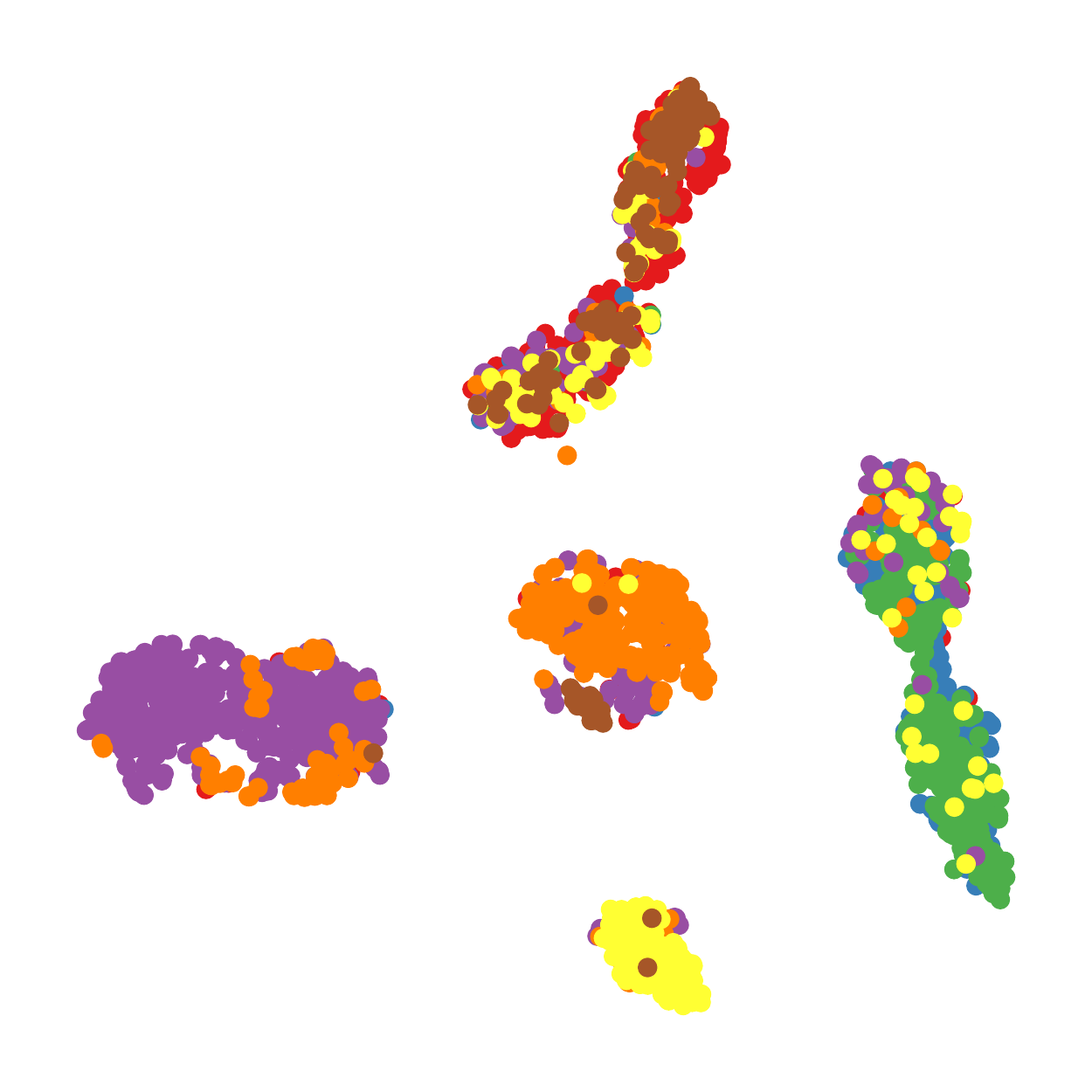}}\hfill
    \subfloat[VCLANC\label{<cora_proposed_tsne>}]{\includegraphics[width=0.1\textwidth]{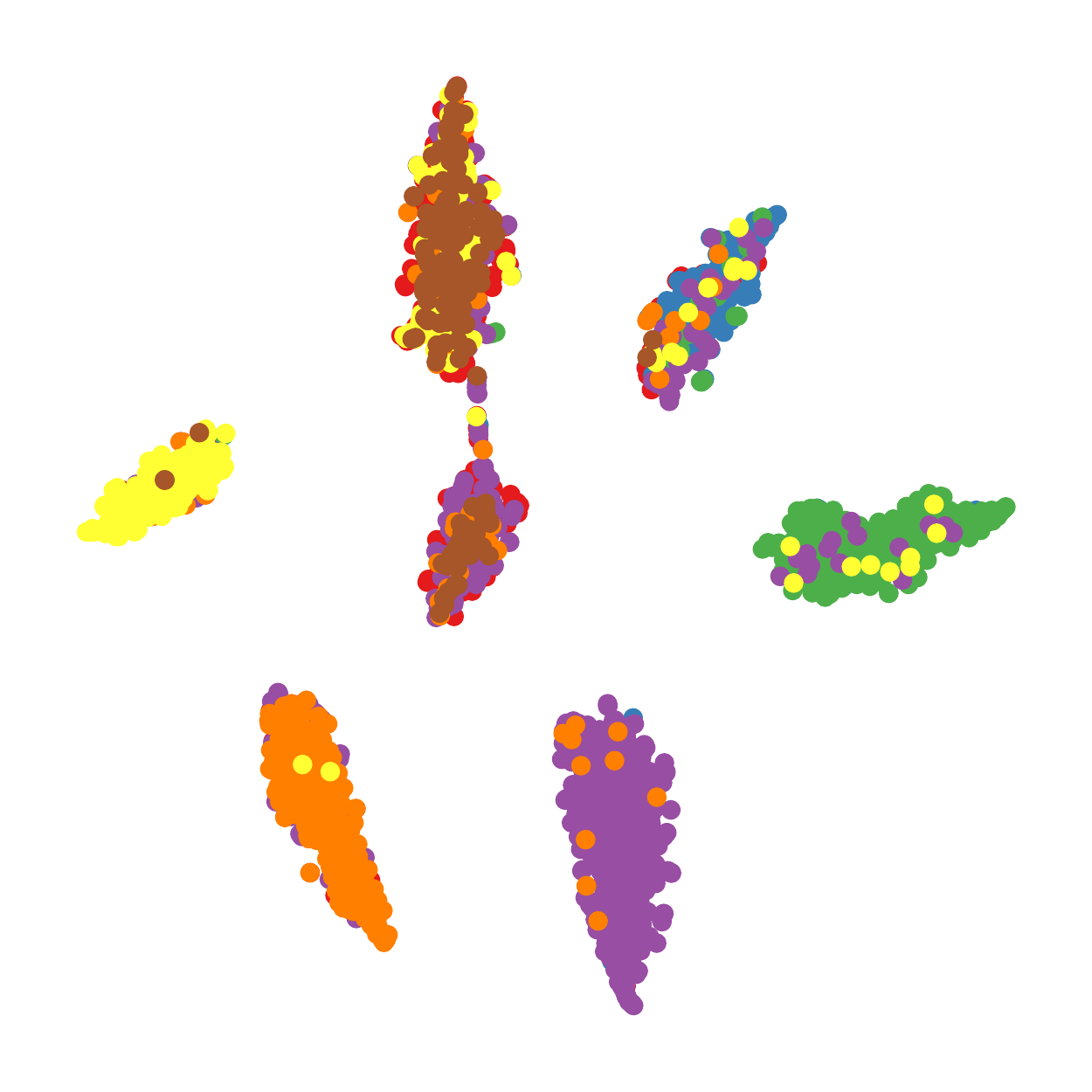}}\hfill 
     \caption{The t-SNE visualization for Cora dataset.}
     \label{fig:tsne_cora}
\end{figure}

To qualitatively analyze the effectiveness of different methods in representation learning, we plot their node embeddings for Cora dataset into a 2-dimensional space using the popular dimension reduction tool: t-SNE \cite{vandermaaten08a}.
Fig. \ref{fig:tsne_cora} presents the visualization results from all the network embedding methods (Fig. \ref{fig:tsne_cora} (b) to Fig. \ref{fig:tsne_cora} (g)) and also the raw attributes of Cora dataset (Fig. \ref{fig:tsne_cora} (a)).
The color of the nodes represents their ground truth categories. 
As we can see in Fig. \ref{fig:tsne_cora} (a), the raw attributes are not separable enough to distinguish different categorical nodes. 
Most nodes from different categories are mixed to each other. 
This drawback is  eased by learning the low-dimensional embeddings for nodes.
From Fig. \ref{fig:tsne_cora} (b) to Fig. \ref{fig:tsne_cora} (d), we show the embedding results of GAE, VGAE and CAN. 
As we can see,  the categorical patterns could be easily noticed and the majority nodes from different categories of nodes are slightly separable to each other.
But the boundary between different node groups are not clear. 
From Fig. \ref{fig:tsne_cora} (e) to Fig. \ref{fig:tsne_cora} (g), we show the embedding results of SDCN, NEC and VCLANC. 
We can see that SDCN does not show good separability for different clusters. 
But NEC and VCLANC have clear boundaries between the different categories. 
The red and brown categories of nodes are hard to be separated by NEC. 
But VCLANC has much pure clusters than NEC. 
For example, the yellow category of nodes is much pure and concentrated in Fig. \ref{fig:tsne_cora} (g) than in Fig. \ref{fig:tsne_cora} (f).
Overall, our proposed method shows much better embedding effectiveness.

\subsection{Running Time Comparison}

\begin{table}[!]
    \centering 
     \caption{Running time comparison (seconds).  }
    \begin{tabular}{c|c|c|c|c}
    \toprule
         Method &  Cora & Citeseer & BlogCatalog & Flickr     \\ \toprule
         GAE&  7.73 &  26.91 & 79.20 & 167.10  \\  
         VGAE& 9.55 & 27.90 &79.31 & 167.66  \\
         CAN& 12.57 & 31.90 & 92.18 & 197.67  \\ 
         SDCN&35.26&42.90 &124.66&198.12       \\
         NEC&19.02&39.89&105.25&220.45    \\ 
         VCLANC&17.75& 36.33 & 97.76 & 204.48 \\
    \bottomrule
    \end{tabular}
    \label{tab:running_time}
\end{table}

We compare the  efficiency of the graph neural network related methods over all the datasets. 
Each method is tested with 300 epochs for fair comparison.  
All the experiments are executed on  an NVIDIA  GPU with 11GB memory.
As shown in Table \ref{tab:running_time}, it is not surprised that GAE shows the least running time among all the methods since GAE only contains the basic encoder and decoder structure. 
For example, on cora dataset, GAE spends 7.73 seconds for running 300 epochs which is the lowest among all the methods.
VGAE is slightly slower than GAE on other dataset as it involves an extra KL divergence loss and ranks as the second most efficient among all the methods.
Its running time on Cora, Citeseer, BlogCatalog and  Flickr is 9.55, 27.90, 79.31 and 167.66, respectively.
Compared with VGAE, CAN is slower since it includes an extra VAE for embedding the attributes.
The running time of CAN on Cora, Citeseer, BlogCatalog and  Flickr is 12.57, 31.90, 92.18 and 197.67, respectively.
Among SDCN, NEC and the proposed VCLANC, SDCN is the most inefficient on  Cora, Citeseer and BlogCatalog datasets. 
This is because that SDCN adopts much deep  graph neural network layers which brings more the parameters. But SDCN shows slighted better efficient than VCLANC on Flickr dataset. 
VCLANC is slightly worse in time consumption than CAN but far better than SDCN.
For example, VCLANC spends 17.75 seconds on Cora dataset while SDCN requires 35.26 seconds and NEC requires 19.02 seconds.
Overall, we can see that VCLANC is more efficient compared with NEC and SDCN for a self-contained clustering task.

\subsection{Embedding Size Analysis}

\begin{figure}[t!]
    \centering
    \subfloat[Cora\label{<cora_embedding>}]{\includegraphics[width=0.2\textwidth]{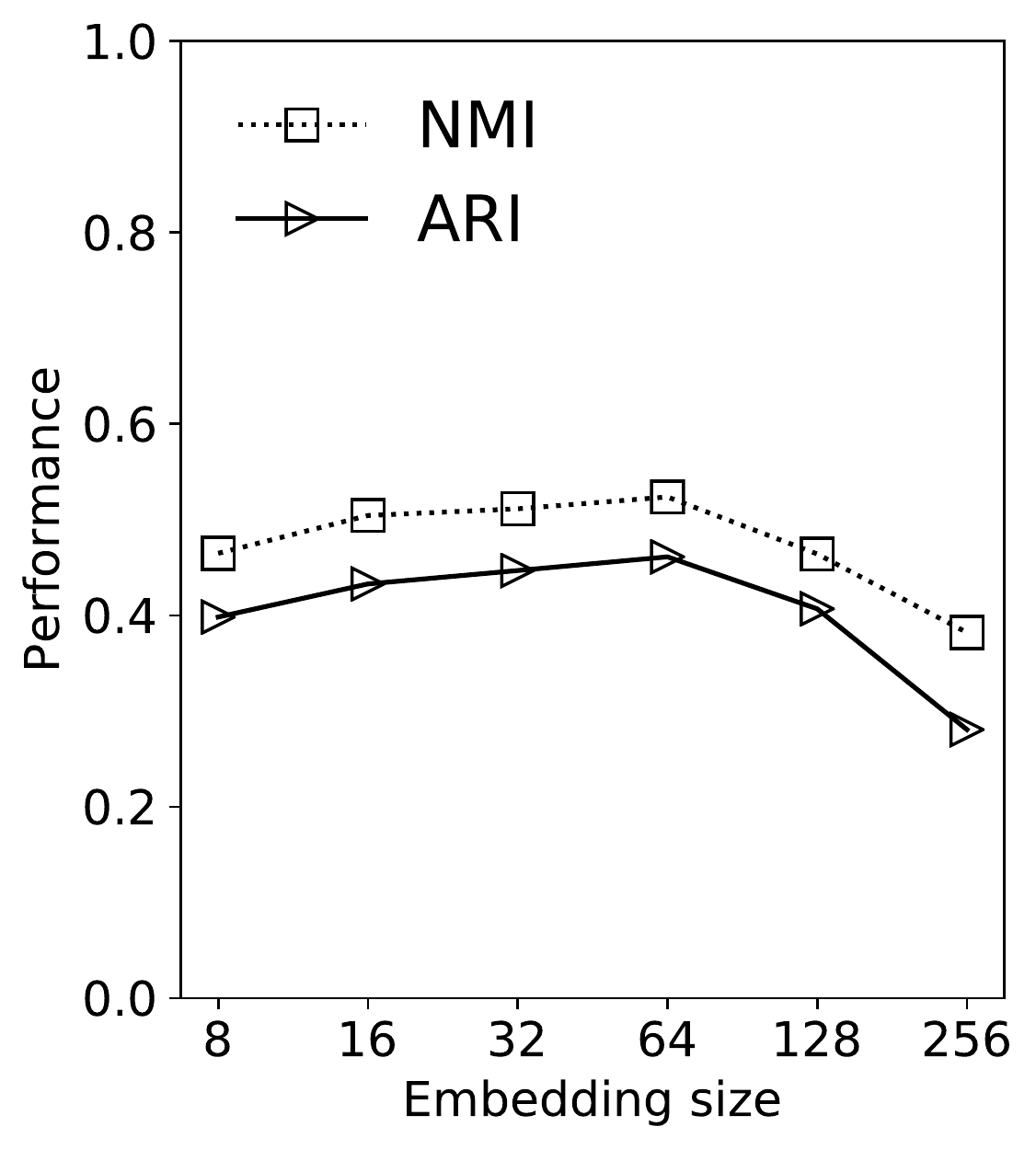}} 
    \subfloat[Citeseer\label{<citeseer_embedding}]{\includegraphics[width=0.2\textwidth]{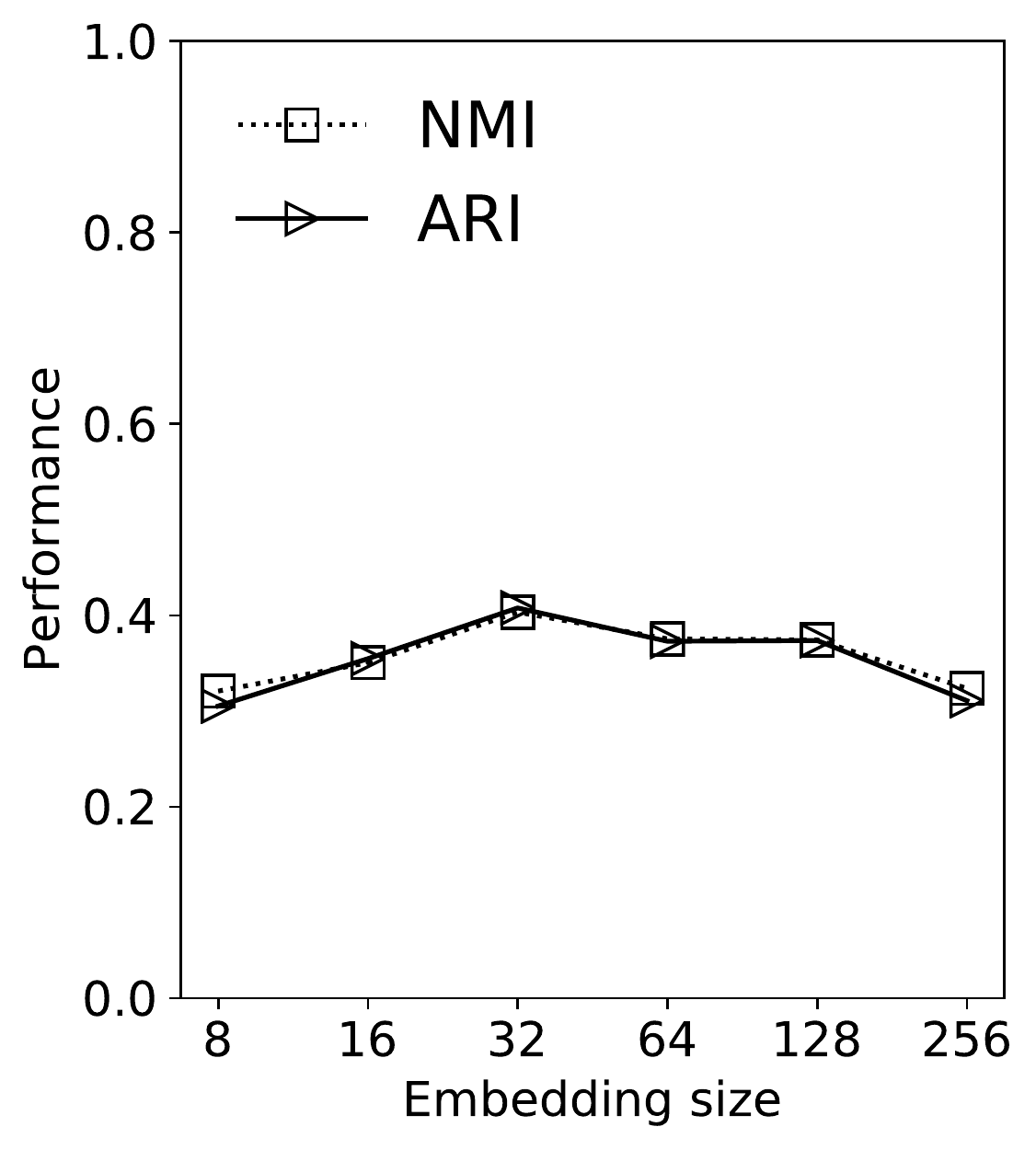}}  \\ 
    \subfloat[Citeseer\label{<blogcatalog_embedding}]{\includegraphics[width=0.2\textwidth]{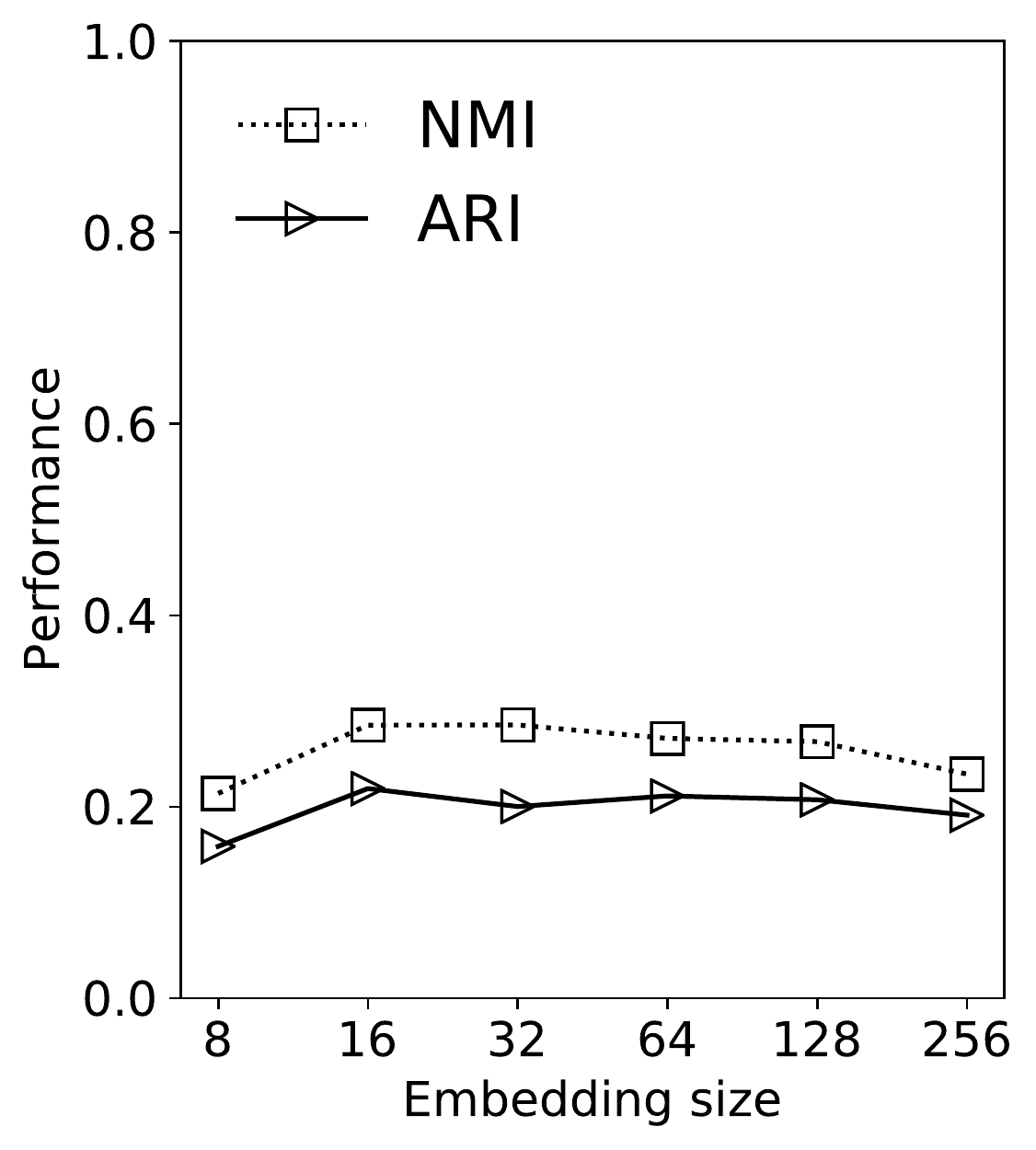}} 
    \subfloat[Flickr\label{<flickr_embedding}]{\includegraphics[width=0.2\textwidth]{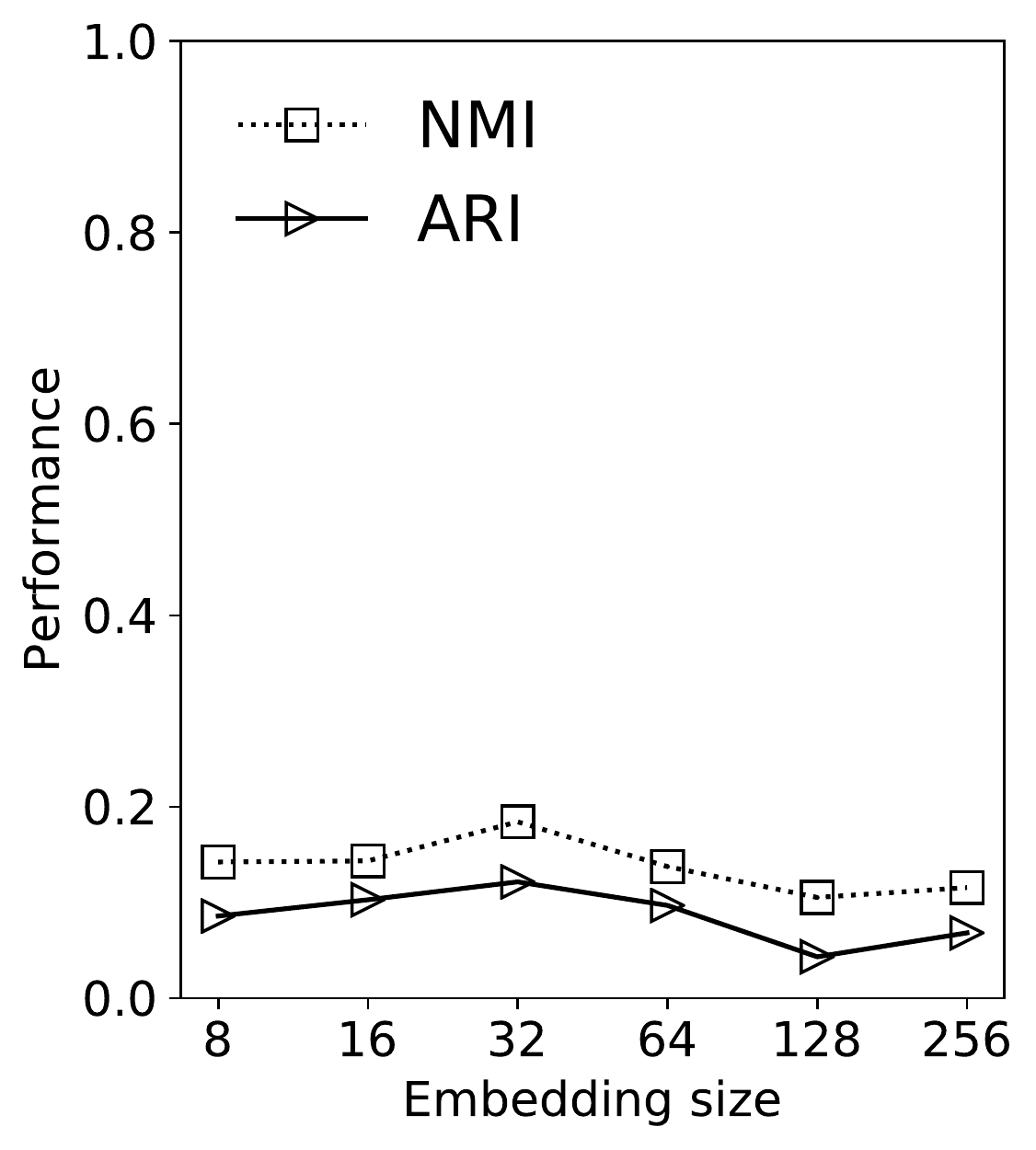}} 
     \caption{Embedding size analysis over all the datasets for the proposed method.}
     \label{fig:embedding_analysis}
\end{figure}

In this section, we analyze the impacts of varying embedding size on the clustering performance for the proposed VCLANC.
Fig. \ref{fig:embedding_analysis} shows the score trend of NMI and ARI by varying the embedding size.
As we can see, when the embedding size is set as 32 or 64, VCLANC could achieve optimal performance on most of the datasets.
Higher or lower embedding size may lead to the feature vectors become too sparse or uninformative to represent the best features for the vertices and thus leads to the decreasing of the clustering performance. The results indicate that VCLANC could learn informative node representations with relatively smaller embedding size.
\section{Conclusions}

In this paper, we have proposed a variational co-embedding learning model for attributed network clustering, called VCLANC.
Compared with the other methods, VCLANC exploits dual variational auto-encoders to embed both nodes and attributes into the same latent space and reconstructs the mutual affinities between nodes and attributes as extra self-supervised knowledge for more distinguished embedding learning.
At the same time, VCLANC adopts a trainable Gaussian mixture priors to infer the nodes clustering assignments in their embedding space.
In this way, the network clustering process is not decoupled from the node representation learning process. 
On the contrary, the embedding space for the nodes could be optimized to find the best partitions for the networks.
We have tested VCLANC on four real-world network datasets to validate its effectiveness. 
The experimental results show that VCLANC outperforms the other methods in regard to different clustering metrics and  learns better node representations based on the visualization analysis.







\bibliographystyle{IEEEtran}
\bibliography{bibfile.bib}

\end{document}